\documentclass[lettersize,journal]{IEEEtran}
\usepackage{array}
\usepackage[caption=false,font=normalsize,labelfont=sf,textfont=sf]{subfig}
\usepackage{textcomp}
\usepackage{verbatim}
\usepackage{hyperref}       

\usepackage{graphics} 
\usepackage{svg}

\usepackage{amssymb}  
\usepackage{textglos}
\usepackage[utf8]{inputenc} 
\usepackage[T1]{fontenc}    
\usepackage{hyperref}       
\usepackage{url}            
\PassOptionsToPackage{hyphens}{url}\usepackage{hyperref}
\usepackage[hyphenbreaks]{breakurl}
\usepackage{booktabs}       
\usepackage{amsfonts}       
\usepackage{nicefrac}       
\usepackage{microtype}      
\usepackage{lipsum}
\usepackage{tabularx} 
\usepackage{amsmath}  
\usepackage{cleveref}
\usepackage{cite} 
\usepackage{titlesec}
\usepackage{mathtools, amssymb, nccmath}
\usepackage{wasysym}
\usepackage{algorithm}
\usepackage{algorithmic}
\usepackage{multicol}

\usepackage{enumitem}
\usepackage{float}
\usepackage{amsmath,amsfonts,amssymb}
\usepackage{cmll}
\usepackage{tensor}
\usepackage{booktabs}
\usepackage{tabularx}
\usepackage{makecell}

\usepackage{tablefootnote}
\usepackage{threeparttable}
\usepackage{multirow}
\usepackage{mathrsfs}  
\usepackage[]{subfig}

\allowdisplaybreaks[1]

\usepackage{setspace}
\usepackage[font=small,skip=5pt]{caption}

\newcolumntype{Z}{ >{\centering\arraybackslash}X }

\newcommand{\crff}{\,\overline{\!\times\!}{}^{\,*}}

\newcommand{\mysp}{.7ex}

\newcommand{\M}{\mathcal{M}}
\newcommand{\F}{\boldsymbol{F}}
\newcommand{\I}{\mathcal{I}}

\newcommand{\T}{^\top}
\newcommand{\Tten}{^{\widetilde{\top}}} 
\newcommand{\Rten}{^{\widetilde{\mathrm{R}}}} 
\newcommand{\RTten}{^{\widetilde{\mathrm{R}},\widetilde{\!\top\!}}} 

\newcommand{\w}{\mathbf{w}}

\newcommand{\N}{\mathbb{N}}

\newcommand{\R}{\mathbb{R}}
\newcommand{\C}{\vC}

\newcommand{\rmvec}[1]{\boldsymbol{#1}}
\newcommand{\greekvec}[1]{\boldsymbol{#1}}

\newcommand{\B}{\boldsymbol{B}{}}
\renewcommand{\C}{\boldsymbol{C}}
\renewcommand{\M}{\boldsymbol{M}}
\newcommand{\A}{\boldsymbol{A}{}}

\renewcommand{\I}{\boldsymbol{I}{}}
\newcommand{\IC}[1]{\I_{#1}^{C}}
\newcommand{\BC}[1]{\B_{#1}^{C}}
\newcommand{\BCT}[1]{\B_{#1}^{C,\top}}

\newcommand{\g}{\rmvec{g}}
\newcommand{\q}{\rmvec{q}}
\renewcommand{\b}{\rmvec{b}}
\renewcommand{\u}{\rmvec{u}}

\newcommand{\U}{\rmvec{U}}
\newcommand{\V}{\rmvec{V}}
\newcommand{\W}{\rmvec{W}}

\usepackage{framed}

\newcommand{\f}{\rmvec{f}}

\renewcommand{\v}{\rmvec{v}}

\renewcommand{\a}{\rmvec{a}}

\newcommand{\qd}{\dot{\q}}
\newcommand{\qdd}{\ddot{\q}}

\newcommand{\Psibar}{\greekvec{\Psi}}
\newcommand{\Psibardot}{\dot{\Psibar}}
\newcommand{\Psibarddot}{\ddot{\Psibar}}

\newcommand{\psibar}{\greekvec{\psi}}
\newcommand{\psibardot}{\,\dot{\!\psibar}}
\newcommand{\psibarddot}{\,\ddot{\!\psibar}}

\newcommand{\taubar}{\greekvec{\tau}}

\newcommand{\red}[1]{{\color{red}#1}}

\newcommand{\blue}[1]{{\color{blue}#1}}

\newcommand{\phibar}{\boldsymbol{s}}
\newcommand{\Phibar}{\boldsymbol{S}}
\newcommand{\phibardot}{\,\dot{\!\phibar}}
\newcommand{\Phibardot}{\,\dot{\!\Phibar}}

\newcommand{\subtree}{\nu}
\newcommand{\subtreeb}{\overline{\nu}}

\usepackage{scalerel}
\newcommand{\timesM}{{\tilde{\smash[t]{\times}}}}
\newcommand{\timesfM}{\timesM {}^*}
\newcommand{\crffM}{\scaleobj{.87}{\,\tilde{\bar{\smash[t]{\!\times\!}}{}}}{}^{\,*}}

\newcommand{\timesf}{{\times}^*}

\newcommand{\dtaudqSO}[3]{\frac{\partial ^{2} \taubar_{#1}}{\partial \q_{#2} \partial \q_{#3}}}

\newcommand{\dtaudqdSO}[3]{\frac{\partial ^{2} \taubar_{#1}}{\partial \qd_{#2} \partial \qd_{#3}}}

\newcommand{\dtaudqdMSO}[3]{\frac{\partial ^{2} \taubar_{#1}}{\partial \q_{#2} \partial \qd_{#3}}}

\newcommand{\dMdq}[3]{\frac{\partial  \M_{#1,#2}}{\partial \q_{#3} }}

\usepackage{tikz}
\usetikzlibrary{matrix,calc}

\setlist[enumerate]{wide=0pt, widest=99,leftmargin=25pt, labelsep=*}

\newcommand{\Ff}{\F}
\newcommand{\Uu}{\U}
\newcommand{\Uv}{\V}

\newcommand{\appref}[1]{App.~\ref{#1}}

\makeatletter
\newlength{\negph@wd}
\DeclareRobustCommand{\negphantom}[1]{%
  \ifmmode
    \mathpalette\negph@math{#1}%
  \else
    \negph@do{#1}%
  \fi
}
\newcommand{\negph@math}[2]{\negph@do{$\m@th#1#2$}}
\newcommand{\negph@do}[1]{%
  \settowidth{\negph@wd}{#1}%
  \hspace*{\negph@wd}
}
\makeatother

\newcommand{\Lie}[2]{\mathscr{L}_{#1} #2}
\newcommand{\X}[2]{X_{#1,#2}} 

\newcommand{\E}{\boldsymbol{E}}
\newcommand{\Tan}{\mathcal{T}}
\newcommand{\SE}{\mathsf{SE}}
\newcommand{\se}{\mathfrak{se}}
\newcommand{\Thomo}{\boldsymbol{T}}
\DeclareMathOperator{\ad}{ad}
\DeclareMathOperator{\Ad}{Ad}
\newcommand{\vf}{\mathbf{f}}
\newcommand{\Xm}[2]{{}^{#1}\! \boldsymbol{X}_{#2}}
\newcommand{\Rm}[2]{{}^{#1}\! \boldsymbol{R}_{#2}}
\newcommand{\Qcal}{\mathcal{Q}}
\renewcommand{\w}{\boldsymbol{w}}

\newcommand{\Bten}[2]{\mathcal{B}(#1,#2)}

\newcommand{\Bicpsiidot}[2]{\mathcal{B}_{#1}^{C}\big(\Psibardot_{#2}\big)}

\newcommand{\Bmat}[2]{\B(#1,#2)}

\newcommand{\parspace}{\vspace{3px}}
\begin{document}

\title{On Second-Order Derivatives of Rigid-Body Dynamics: Theory \& Implementation}

\author{Shubham Singh,~\IEEEmembership{Member,~IEEE,} Ryan P. Russell~\IEEEmembership{Member,~IEEE,} and Patrick M.~Wensing,~\IEEEmembership{Member,~IEEE}
\thanks{Shubham Singh and Ryan P. Russell are with the Dept. of Aerospace Engineering, The University of Texas at Austin, TX-78751, USA. \href{mailto:singh281@utexas.edu}{singh281@utexas.edu}, \href{mailto:ryan.russell@utexas.edu}{ryan.russell@utexas.edu}}
\thanks{Patrick M. Wensing is with the Dept. of Aerospace \& Mechanical Engineering, University of Notre Dame, IN-46556, USA. \href{mailto:pwensing@nd.edu}{pwensing@nd.edu}
       } 
    \thanks{This work was supported in part by the National Science Foundation grants CMMI-1835013 and CMMI-1835186.}
}

\markboth{Journal of \LaTeX\ Class Files,~Vol.~14, No.~8, August~2021}%
{Shell \MakeLowercase{\textit{et al.}}: A Sample Article Using IEEEtran.cls for IEEE Journals}


\maketitle


\begin{abstract}
Model-based control for robots has increasingly been dependent on optimization-based methods like Differential Dynamic Programming and iterative LQR (iLQR). These methods can form the basis of Model-Predictive Control (MPC), which is commonly used for controlling legged robots. Computing the partial derivatives of the dynamics is often the most expensive part of these algorithms, regardless of whether analytical methods, Finite Difference, Automatic Differentiation (AD), or Chain-Rule accumulation is used. Since the second-order derivatives of dynamics result in tensor computations, they are often ignored, leading to the use of iLQR, instead of the full second-order DDP method. In this paper, we present analytical methods to compute the second-order derivatives of inverse and forward dynamics for open-chain rigid-body systems with multi-DoF joints and fixed/floating bases. An extensive comparison of accuracy and run-time performance with AD and other methods is provided, including the consideration of code-generation techniques in C/C++ to speed up the computations. For the 36 DoF ATLAS humanoid, the second-order Inverse, and the Forward dynamics derivatives take $\approx 200 \mu s$, and $\approx 2.1  ms$ respectively, resulting in a $3 \times$ speedup over the AD approach. 

\end{abstract}



\section{Introduction}
In recent years, optimization-based methods have become popular for robot motion generation and control. Although full second-order (SO) optimization methods offer superior convergence properties, most work has focused on using only the first-order (FO) dynamics approximation, such as in iLQR~\cite{tassa2012synthesis, koenemann2015whole,dantec2021whole}.
Differential Dynamic Programming (DDP)~\cite{mayne1966second} is a use-case for full SO optimization that has gained wide interest for robotics applications~\cite{tassa2014control,chatzinikolaidis2021trajectory,mastalli2020crocoddyl,li2020hybrid}, with many variants of DDP using multiple shooting~\cite{pellegrini2020multiple} and parallelization~\cite{plancher2018performance}  developed for computational and numerical improvements. Beyond DDP, the SO derivatives of dynamics may also be useful for several other optimization algorithms, like Sequential-Quadratic Programming (SQP)~\cite{wright1999numerical}, commonly used for trajectory optimization more broadly.

The state-of-the-art for including SO dynamics derivatives in trajectory optimization is the work by Lee et al.~\cite{lee2005newton}, where they use forward chain-rule expressions and recursive expressions to calculate the SO partial derivatives of joint torques with respect to joint configuration and joint rates for revolute and prismatic joint models. 
Nganga and Wensing~\cite{nganga2021accelerating} presented a method for getting the SO directional derivatives of inverse dynamics as needed in the backward pass of DDP by employing reverse mode Automatic Differentiation (AD) and a modified version of the Recursive Newton Euler Algorithm (RNEA)~\cite{Featherstone08}. Although this strategy avoids the need for the full SO partial derivatives of RNEA, it lacks the opportunity for parallel computation across the trajectory. Full SO partial derivatives calculation, on the other hand, can be easily parallelized to accelerate the backward pass of DDP or other SQP methods. 

There are many strategies that can be considered for computing derivatives. In the past, finite-difference methods were used to obtain the derivatives of Rigid-Body Dynamics (RBD) for trajectory optimization~\cite{tassa2012synthesis,koenemann2015whole}. The finite-difference approach can be parallelized, but requires tuning and still suffers from low accuracy. Inaccurate Jacobian and Hessian approximations during optimization often lead to ill-conditioning, and poor convergence~\cite{lee2005newton}. The complex-step method~\cite{squire1998using,lantoine2012using,cossette2020complex} is an alternate technique to obtain derivatives of any function or algorithm, while maintaining accuracy to machine precision. However, the method suffers from high computation time due to the reliance on complex arithmetic.

Automatic Differentiation (AD) or Algorithmic Differentiation is a popular technique to obtain the derivatives of a function or an algorithm. AD methods are often categorized into two main types a) source-code transformation, where the semantics of the code is re-written explicitly to include the computation of the derivatives along with the function \cite{hascoet2013tapenade} or b) Operator-overloading techniques, which depend on re-defining elementary operators used in a function or algorithm to propagate additional information during computation. Sophisticated AD tools like CasADi~\cite{andersson2012casadi} and ADOL-C~\cite{walther2009getting} use operator overloading strategies to build an expression graph of the function to be differentiated. This approach supports both forward-mode and reverse-mode sensitivity propagation. A simpler AD approach uses operator overloading without explicitly building an expression graph. Instead, sensitivity information is propagated along the original computational flow of the algorithm.  As a result, this second strategy is limited to forward-mode AD. The library COSY \cite{makino2006cosy} is an example of this category. 


AD tools have been used in the past in several Rigid-Body Dynamics (RBD) libraries for motion planning. The library Drake~\cite{tedrake2019drake} uses AD for computing FO derivatives of the dynamics. RobCoGen~\cite{giftthaler2017automatic} is based on using CppAD~\cite{bell2012cppad} along with code generation techniques to get the FO derivatives of Inverse and Forward Dynamics. The open-source symbolic AD toolbox CasADi~\cite{andersson2012casadi} has been a popular choice in the past for RBD libraries~\cite{johannessen2019robot,astudillo2021mixed} and computing the derivatives of dynamics. CasADi employs forward and reverse mode chain-rule differentiation, and also supports code generation to output compilable C code. 

An alternative to relying on general-purpose AD tools is to analytically derive the desired derivatives and then develop special-purpose numerical methods for them. 
The FO analytical derivatives in Ref.~\cite{carpentier2018analytical} were directly compared with RobCoGen's AD derivatives, showing significant performance improvements via the analytical approach. Kudruss et al.~\cite{kudruss2019efficient} also showed the benefits of using analytical derivatives that exploit the structure of RBD algorithms over naively using the AD approach. Astudillo et al.~\cite{astudillo2021mixed} used CasADi and code generation techniques to get the FO and SO derivatives of Inverse and Forward Dynamics. However, they also concluded that, for the FO derivatives,  using analytical derivatives leads to a reduction in run-time over the AD approach. Although most optimization pipelines use the derivatives of the Forward Dynamics, the latest work by Mastalli et al.~\cite{mastalli2022inverse} shows the computational benefits of using Inverse Dynamics constraints in an optimal control framework, pushing the need for Inverse Dynamics derivatives as well.


Manually taking analytical derivatives is often neglected for systems with highly non-linear dynamics like RBD due to the tediousness and time-consuming nature of the derivation process. Although significant work has been done for deriving FO analytical partial derivatives of inverse/forward dynamics~\cite{jain1993linearization,ayusawa2018comprehensive,carpentier2018analytical,singh2022efficient,bos2022}, the literature still lacks analytical SO partial derivatives of rigid-body dynamics. This omission is mainly due to the tensor nature of SO derivatives, and the lack of established tools for working with dynamics tensors. Hence, in this work, we provide a tensorial framework, followed by simplified analytical expressions for the SO derivatives of both Inverse and Forward dynamics.

 \subsection*{Contributions}
 This paper builds on the previous conference paper~\cite{singh2022analytical}, where the authors presented the analytical second-order derivatives of Inverse Dynamics w.r.t. the joint configuration ($\q$),  velocity vector ($\qd$), and joint acceleration ($\qdd$) 
 using the tensorial extensions of Spatial Vector Algebra (SVA). These developments are reviewed in Sec.~\ref{sec:SVA_tensor} and~\ref{sec:ID_SO_theory}, with more details for clarity. Beyond the conference paper, we first study the benefits of the Inverse Dynamics SO derivatives over the state-of-the-art methods like Chain-Rule accumulation and Automatic Differentiation via extensive comparison in terms of accuracy and run-times (Sec.~\ref{sec:ID_SO_results}). 
 Following that, we detail new efficient techniques to compute the analytical SO derivatives of Forward Dynamics by exploiting the FO/SO derivatives of Inverse Dynamics  (Sec.~\ref{sec:FD_SO_theory}). Finally, a run-time and accuracy comparison for the analytical SO Forward Dynamics derivatives with the state-of-the-art is presented in Sec.~\ref{sec:FD_SO_results}. Overall, the paper contributes a comprehensive development and numerical evaluation of algorithms for the SO derivatives for RBD.


\section{Rigid-Body Dynamics Background}
\label{sec:pd_rbd}

\noindent {\bf Rigid-Body Dynamics}: 
We consider an open-chain rigid-body system with $N$ joints and $n$ total degrees of freedom (DoFs). We partition the configuration manifold $\Qcal$ as $\Qcal = \Qcal_{1} \times ... \times \Qcal_{N}$, where each $\Qcal_{i}$ represents the configuration manifold of joint $i$. Here, each $\Qcal_i$ is assumed to be a sub-group of $\SE(3)$, which allows the modeling of revolute, prismatic, helical, spherical, and floating-base joints. The number DoFs of joint $i$ is denoted by $n_{i} \in \N_{+}$, and $n=\sum_{i=1}^{N} n_i$. The state variables associated with each joint are the configuration $\q_{i} \in \Qcal_{i}$ and generalized velocity vector $\qd_i\in\mathbb{R}^{n_{i}}$, while the control variable is the generalized force/torque vector $\taubar_{i} \in \mathbb{R}^{n_{i}}$. With this convention, $\qd_i$ uniquely specifies the time rate of change for $\q_i$ without strictly being its time derivative. The Inverse Dynamics (ID), is given by 
 \begin{align}
    \taubar &= \M(\q)\qdd+ \C(\q,\qd)\qd + \g(\q)
    \label{inv_dyn} \\
&= \textrm{ID}({\rm model},\q,  \qd, \qdd)
    \label{inv_dyn_model}
\end{align}
where $\M \in \R^{n \times n}$ is the mass matrix, $\C \in \R^{n \times n}$ is the Coriolis matrix, and  $\g \in \R^{n}$ is the vector of generalized gravitational forces. For a fixed configuration and velocity, ID computes $\taubar$ given $\qdd$, while Forward Dynamics (FD) computes $\qdd$ for a given $\taubar$:
\begin{align}
    \qdd &= \M^{-1}(\q)\left( \taubar -  \C(\q,\qd)\qd - \g(\q) \right)
    \label{fwd_dyn2} \\
    &= \textrm{FD}({\rm model},\q, \qd, \taubar)
    \label{for_dyn_model}
\end{align} 
Efficient $\mathcal{O}(N)$ algorithms for ID and FD are, respectively, the Recursive-Newton-Euler-Algorithm (RNEA) \cite{orin1979kinematic,Featherstone08} and  the Articulated-Body-Algorithm (ABA)~\cite{vereshchagin1974computer,featherstone1983calculation,Featherstone08}.

 
\parspace
\noindent {\bf Notation:} In this paper, Cartesian vectors are denoted with lower-case letters with a bar ($\bar{v}$), spatial (6D) vectors \cite{Featherstone08} with lower-case bold letters (e.g., $\a$), matrices with capitalized bold letters (e.g., $\boldsymbol{A}$), and tensors with capitalized calligraphic letters (e.g., $\mathcal{A}$). As a matter of convention, $n$-vectors $\qd, \qdd, \taubar$ are also denoted by lower-case bold letters. 

Spatial vectors are 6D vectors that combine the linear and angular aspects of a rigid-body motion or net force~\cite{Featherstone08}. Spatial motion vectors, such as velocity and acceleration, belong to a 6D vector space denoted $M^{6}$. Force-like vectors, such as force and momentum, belong to another 6D vector space $F^{6}$ \cite{Featherstone08}. Spatial vectors are usually expressed in either the ground coordinate frame or a body coordinate (local) frame. For example, the spatial velocity ${}^{k}\v_{k} \in M^{6}$ of a body $k$ expressed in the body frame is
\begin{align}
{}^{k}\v_{k} = \begin{bmatrix}
           {}^{k}\bar{\omega}_{k} \\
           {}^{k}\bar{v}_{k}
         \end{bmatrix}
\end{align}
where ${}^{k}\bar{\omega}_{k}$ $\in \mathbb{R}^{3}$ is the angular velocity expressed in a coordinate frame fixed to the body, while ${}^{k}\bar{v}_{k}$ $\in \mathbb{R}^{3}$ is the linear velocity of the origin of the body frame. 
We can likewise express the spatial velocity in an earth-fixed ground frame (labeled 0) as
\[
{}^{0}\v_{k} = \underbrace{\begin{bmatrix} \Rm{0}{k} & \boldsymbol{0}_{3\times 3} \\  ({}^0 \boldsymbol{p}_{k} \times) \Rm{0}{k} & \Rm{0}{k} \end{bmatrix}}_{ \Xm{0}{k}} {}^k \v_k
\]
where $\Xm{0}{k}$ is a spatial transform \cite{Featherstone08}, $\Rm{0}{k}$ is a rotation matrix from frame $k$ to $0$, ${}^0 \boldsymbol{p}_{k}$ is the vector to the origin of frame $k$, and $(\boldsymbol{p} \times)$ denotes the Cartesian cross-product matrix.
When the frame used to express a spatial vector is omitted (e.g., $\v_k$), the ground frame is assumed.

 A spatial cross product between motion vectors $\v$ and $\u$, written as $(\v \times) \u$, is given by \eqref{cross_defn}. This operation gives the time rate of change of $\u$, when $\u$ is moving with a spatial velocity $\v$.  
A spatial cross product between a motion vector $\v$ and a force vector $\f$ is written as $(\v \times^*) \f$, as defined by \eqref{cross_f_defn}.
\begin{multicols}{2}
  \begin{align}
    \v \times = \begin{bmatrix}
       \bar{\omega}  \times & \bf{0}  \\
       \bar{v} \times &  \bar{\omega} \times
    \end{bmatrix}
    \label{cross_defn}
\end{align}
  \begin{align}
    \v \times^* = \begin{bmatrix}
        \bar{\omega} \times & \bar{v} \times  \\
        \bf{0} &  \bar{\omega} \times
    \end{bmatrix}
    \label{cross_f_defn}
\end{align}
\end{multicols}
\noindent An operator $\crff$ is defined by swapping the order of the cross product, such that $(\f \crff) \v = (\v \times^*) \f$~\cite{echeandia2021numerical}.  Further introduction to  SVA is provided in Ref.~\cite{Featherstone08}. 

\begin{table}[]
\small
    \centering
    \begin{tabular}{c c  c c}
    \hline
         \multicolumn{2}{c }{\bf Spatial Vector Algebra} &  \multicolumn{2}{c}{\bf Lie-Theoretic}  \\ 
         {\em Nomenclature} & {\em Notation}   &  {\em Nomenclature} & {\em Notation} \\ \hline 
         Spatial Transform & $\Xm{0}{i}$ & Adjoint  & $\Ad_{{}^0 \Thomo_i}$ \\
         Motion Cross Product & $(\u \times)$ & adjoint  & $\ad_{\u}$ \\
         Force Cross Product & $(\u \times^*)$ & coadjoint  & $-\ad_{\u}^*$  \\ \hline
    \end{tabular}
    \caption{Notation/nomenclature conversion between spatial vector algebra notation and that of a Lie-theoretic treatment of dynamics. ${}^0 \Thomo_i \in \SE(3)$ denotes a homogenous transform.}
    \label{tab:correspondence}
\end{table}
Spatial vector algebra is closely connected with a Lie-theoretic treatment of dynamics \cite{park2018geometric}. We make a few notational connections explicitly in Tab.~\ref{tab:correspondence} and refer the reader to \cite{traversaro2019multibody} for additional detail. 
Given a spatial motion vector $\v$, we employ the usual \textit{hat} operator~\cite[Sec. 3.2]{murray2017mathematical} ($\wedge$) to return an element in the Lie algebra $\se(3)$, while the \textit{vee} operator ($\vee$) reduces a Lie algebra element to its components:
    \begin{align}
        \v^\wedge &= \begin{bmatrix} \bar{\omega}\times & \bar{v}   \\
    \boldsymbol{0}_{1 \times 3} & 0
    \end{bmatrix} & 
         \begin{bmatrix} \bar{\omega}\times & \bar{v}   \\
    \boldsymbol{0}_{1 \times 3} & 0
    \end{bmatrix}^{\vee} &= \v
    \end{align}

\begin{figure}[tb]
 \centering
 \includegraphics[width=.80 \columnwidth]{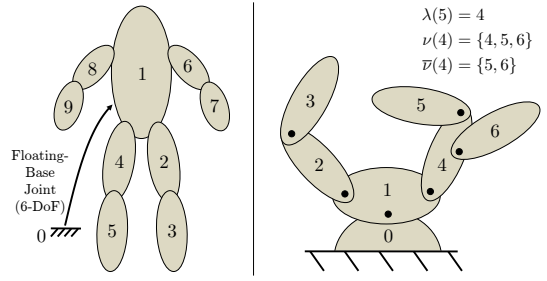}
 \caption{Convention examples with floating- and fixed-base systems showing parent-child, sub-tree and successor nomenclature.}
 \label{link_fig}
 \end{figure}

\parspace
\noindent {\bf Connectivity:} An open-chain kinematic tree (Fig.~\ref{link_fig}) is considered with the $N$ links connected by $N$ joints, each with up to 6 DoF. Body $i$'s parent toward the root of the tree is denoted as $\lambda(i)$, 
and we define $i \preceq j$ if body $i$ is in the path from body $j$ to the root.  Joint $i$ is defined as the connection between body $i$ and its predecessor. The set of bodies in the subtree rooted at body $i$ is denoted as $\subtree(i)$, while $\subtreeb(i)$ denotes the set of bodies in $\subtree(i)$, excluding the body $i$.

As mentioned before, we consider joints whose configurations form a sub-group of the Lie group $\SE(3)$, and we make the following choices for generalized velocities. For a prismatic joint, the configuration $\q_{i} \in \Qcal_i \cong \mathbb{R}$, while $\qd_{i}\in \mathbb{R}$  gives the linear translation rate. For a revolute joint, $\q_{i} \in \Qcal_i \cong SO(2)$, and $\qd_{i} \in \mathbb{R}$ gives the rotational rate of the joint. For a spherical joint, $\q_{i} \in \Qcal_i \cong SO(3)$, and $\qd_{i} = {}^i \bar{\omega}_{i/\lambda(i)} \in \R^3$ gives relative angular velocity between neighboring bodies. 
For a 6-DoF free motion joint, $\q_{i} \in  \SE(3)$, and $\qd_{i} = {}^i \v_{i/\lambda(i)} \in \R^6$. While we can use compact configuration representations for implementation (e.g., using a scalar to store $\q_i$ for a prismatic joint), we emphasize that herein we will identify each $\q_i$ with its corresponding homogeneous transform in $\Qcal_i \subseteq \SE(3)$ to ease the exposition. 

The spatial velocities of the neighboring bodies in the tree are then related by the recursive expression
\[
\v_{i} = \v_{\lambda(i)} + \Phibar_{i} \qd_{i}\,,
\]
where $\Phibar_{i} \in \mathbb{R}^{6\times n_i}$ is the joint motion subspace matrix for joint $i$ \cite{Featherstone08}. 
 The velocity $\v_{i}$ can also be written in an analytical 
 form as the sum of joint velocities over predecessors as $\v_{i} = \sum_{l \preceq i} \Phibar_{l} \qd_{l}$.
 The derivative of the joint motion subspace matrix in local coordinates (often denoted $\,\mathring{\!\Phibar}_{i}$~\cite{Featherstone08}) is assumed to be zero. As such, we can express $\Phibar_i$ in local coordinates as a fixed matrix:
 \[
{}^i\Phibar_i = \begin{bmatrix} {}^i \phibar_{i,1} & \cdots & {}^i\phibar_{i,n_i} \end{bmatrix}
 \]
For later use, the quantity $\Phibardot{}_{i} = \v_{i} \times \Phibar_{i}$ signifies the rate of change of $\Phibar_{i}$ due to the local coordinate system moving.

\parspace
\noindent {\bf Dynamics:} The spatial equation of motion~\cite{Featherstone08} is given for body $k$ as 
$\f_{k}= \I_{k}\a_{k} + \v_{k} \times^*\I_{k} \v_{k}$, 
where $\f_{k}$ is the net spatial force on body $k$, $\I_{k}$ is its spatial inertia~\cite{Featherstone08}, and $\a_{k}$ is its spatial acceleration. The inverse dynamics of the mechanism can be written as
\[
\taubar_i = \Phibar_{i}\T \f{}_{i}^{C}
\]
where $\taubar_{i}$ represents the joint torques/forces for joint $i$ from \eqref{inv_dyn}, and $\f{}_{i}^{C} = \sum_{k \succeq i} \f_{k}$ is the composite spatial force transmitted across joint $i$. For later development, we likewise consider the composite rigid-body inertia of the sub-tree rooted at body $i$, given as $\I{}_{i}^{C} = \sum_{k \succeq i} \I_{k}$, which represents the locked inertia of all bodies in the subtree.


\parspace
\noindent {\bf Notation for FO Derivatives:}\footnote{This section is developed slowly in response to reviewers of previous papers commenting that the exact nature of our derivatives was unclear. Any feedback here would be particularly appreciated.}
We will consider the derivatives of the ID and FD with respect to configuration changes of each joint. We do so by considering a collection of Lie derivatives for each manifold $\Qcal_i$ as follows. 

For any vector field $X$ on $\Qcal$ and any scalar function $f:\Qcal\rightarrow \R$, we denote by $\Lie{X}{f}:\Qcal\rightarrow \R$ the Lie derivative of $f$ along $X$ as:
\begin{equation}
\Lie{X}{f}(\q) = \lim_{t \rightarrow 0} \frac{ f( \varphi_t^X(\q)) - f(\q)}{t}
\label{lie_deriv_defn}
\end{equation}
where $\varphi_t^X(\q)$ gives the result of flowing along the vector field $X$ for $t$ seconds when starting from $\q$\cite{murray2017mathematical}. 
For any vector-valued function $\vf:\Qcal\rightarrow \R^m$ we let $\Lie{X}{\vf}$ give the component-wise Lie derivatives. 
%
We denote $X( \q) \in \Tan_{\q} \Qcal$ as the evaluation of the vector field at $\q \in \Qcal$ where $\Tan_{\q} \Qcal$ denotes the tangent space to $\Qcal$  at $\q$. In this regard, $\Lie{X}{\vf}(\q)$ can be interpreted as  the directional derivative of $\vf$ at $\q$ in the direction $X(\q)$. For example, if $X$ is selected to correspond to unit-rate rotations of a revolute joint on the mechanism, then $\Lie{X}{\vf}$ represents a conventional partial derivative of $\vf$ w.r.t. the joint angle.

To specify the directions of all possible configuration changes at each joint, we consider a set of basis vectors fields $\X{i}{1}, \ldots, \X{i}{n_i} $, i.e., such that at each $\q_i \in \Qcal_i$
\[
{\rm span}( \X{i}{1}(\q_i) , \ldots, \X{i}{n_i}(\q_i) ) = \Tan_{\q_i} \Qcal_i 
\]
Since $\Qcal_i$ is a subgroup of $\SE(3)$ we construct such a basis via a fixed set of generators $\E_{i,1},\ldots, \E_{i,n_i} \in \se(3)\subset \R^{4 \times 4}$ that we select according to $[\E_{i,j}]^\vee =  {}^i \phibar_{i,j}$. With this construction we select basis vector fields $\X{i}{j}$ such that for any $\Thomo \in \Qcal_i \subseteq \SE(3)$
\begin{equation}
\X{i}{j}( \Thomo ) = \Thomo \, \E_{i,j} \in \Tan_{ \Thomo } \Qcal_i
\label{eq:field_defns}
\end{equation}

We can identify each of these vector fields $\X{i}{j}$ on $\Qcal_i$ as a vector field on $\Qcal$ where only joint $i$ is moving. We thus consider the following set of Lie derivatives of the function $\vf:\Qcal \rightarrow \mathbb{R}^m$ as the first-order ``partial derivatives w.r.t. $\q_i \in \Qcal_i$'' as:
\begin{equation}
    \Lie{\q_i}{\vf} \triangleq \begin{bmatrix} \Lie{\X{i}{1}}{\vf} & \cdots & \Lie{\X{i}{n_i}}{\vf} \end{bmatrix}
    \label{multi_dof_lie_deriv}
\end{equation}

If we suppose that $\vf:\Qcal_i \rightarrow \R$, then each of these Lie derivatives is equivalently given as:
\[
\Lie{\X{i}{j}}{\vf}( \Thomo )  = \lim_{\epsilon\rightarrow 0} \frac{ \vf( \Thomo {\rm exp}(\E_{i,j} \epsilon) ) - \vf(\Thomo) }{\epsilon}
\]
where $\epsilon$ is a perturbation parameter\footnote{These derivatives also called {\em right Lie derivatives} by Chirikjian \cite{chirikjian2011stochastic} and denoted as $\Lie{\E_{i,j}}{^r \f}$ therein since the perturbation is applied on the right.}. Together for all $\q \in \Qcal$ a similar operation is denoted as $\Lie{\q}{\vf}$, which we take the notational liberty of denoting as $\frac{\partial}{\partial \q} \vf $.

As a relevant example, let us consider the FO partial derivatives of $\taubar = {\rm ID}({\rm model}, \q,\qd,\qdd)$ \eqref{inv_dyn_model} which we view as a function ${\rm ID} : \Qcal\times \R^n \times \R^n \rightarrow \R^n$. 
The operator $\frac{\partial}{\partial \q_k}$ represents a collection of Lie derivatives where $\frac{\partial \taubar_{i}}{ \partial \q_k}$ is the $n_i \times n_k$ matrix with columns giving the derivatives of $\taubar_{i} \in \mathbb{R}^{n_i}$ w.r.t~changes in configuration along each of the $n_k$ free modes of joint $k$.


\parspace
\noindent {\bf FO Dynamics Derivatives:}  Previous work \cite{singh2022efficient} developed analytical expressions for the FO partial derivatives of ID w.r.t. $\q$ and $\qd$, with other related formulations in~\cite{jain1993linearization, ayusawa2018comprehensive}. Summarizing the main results:
\begin{subequations}
 \label{dtau_dq}
\begin{eqnarray}
     \frac{\partial {\taubar_{i}}}{\partial \q_{j}} &=&  { \Phibar_{i}\T \big[ 2  \B_{i}^{C} \big] \Psibardot{}_{j} + \Phibar_{i}\T  \I_{i}^{C} \Psibarddot{}_{j} }, (j \preceq i)
          \label{dtau_dq_1} 
\\
\frac{\partial {\taubar_{j}}}{\partial {\q_{i}}} &=&  {\Phibar_{j}\T [ 2 \B_{i}^{C} \Psibardot_{i} +  \I_{i}^{C}  \Psibarddot_{i}+(\f_{i}^{C} )\crff \Phibar_{i}  ], (j \prec i) }~~ 
     \label{dtau_dq_2}
\end{eqnarray}
\end{subequations}
\vspace{-10px}
\begin{subequations}
\label{dtau_dqd}
\begin{eqnarray}
         \frac{\partial {\taubar_{i}}}{\partial {\dot{\q}_{j}}} &=& {\Phibar_{i}\T \Big[2 \B_{i}^{C} \Phibar_{j} +  \I_{i}^{C} (    \Psibardot_{j} + \Phibardot_{j} ) \Big],  (j \preceq i) }
         \label{dtau_dqd_1}
\\
    \frac{\partial {\taubar_{j}}}{\partial {\dot{\q}_{i}}}  &=& {\Phibar_{j}\T \Big[2 \B_{i}^{C} \Phibar_{i} +  \I_{i}^{C} (    \Psibardot_{i} + \Phibardot_{i} ) \Big], (j \prec i)}~~~~~~~~\,
         \label{dtau_dqd_2}
     \end{eqnarray}
    \end{subequations}
where the quantity $\B_{k}\triangleq \Bmat{\I_{k}}{\v_{k}}$ is a body-Coriolis matrix \cite{singh2022efficient,echeandia2021numerical},
\begin{equation}
\Bmat{\I_{k}}{\v_{k}} = \scaleobj{.75}{ \frac{1}{2}} [(\v_k \times^* )\I_{k} - \I_{k} (\v_k \times) + (\I_{k} \v_k ) \crff]
\label{bk_term_defn}
\end{equation}
while $\BC{i}$ is its composite given by $\B{}_i^C = \sum_{k\succeq i} \B_k$. 
 The quantities $\Psibardot_{j}$ and $\Psibarddot_{j}$ represent the time-derivative of $\Phibar_{j}$, and $\Psibardot_{j}$, respectively, due to joint $j$'s predecessor $\lambda(j)$ moving.
\begin{align}
\begin{split}
  \Psibardot_{j} = & \v_{\lambda(j)} \times \Phibar_j \\
 \Psibarddot_{j} = & \a_{\lambda(j)} \times \Phibar_{j} + \v_{\lambda(j)} \times \Psibardot_{j}
\end{split}
\label{psidot_psidotdot_defn}
\end{align}

\parspace
\noindent {\bf Notation for SO Derivatives:} For a scalar-valued function $f :\Qcal \rightarrow \R$, we denote $\nabla_{\q} f \triangleq [ \Lie{\q}{f}]\T: \Qcal \rightarrow \R^n$. Subsequently, the Lie derivatives of the vector-valued function $\nabla_{\q} f$ are:
\begin{equation}
\Lie{\q}{} \left[  \nabla_{\q} f  \right] = \begin{bmatrix} \Lie{X_1}{ \Lie{X_1}{f}} & \cdots  & \Lie{X_n}{ \Lie{X_1}{f}} \\
\vdots & \ddots & \vdots \\
\Lie{X_1}{ \Lie{X_n}{f}} & \cdots & \Lie{X_n}{ \Lie{X_n}{f}}
\end{bmatrix}
\label{SO_lie_deriv}
\end{equation}
where $X_1, ..., X_n$ are  $X_{1,1}$,...., $X_{1,n_1}$,...., $X_{N,1}$,...., $X_{N, n_N}$ relabeled in order for simplicity. With again a liberal choice of notation, the above type of operation is denoted as:
\begin{align}
\frac{\partial^2 f}{\partial \q_a \partial \q_b} = \Lie{\q_b} \left[ \nabla_{\q_a} f \right] : \Qcal \rightarrow \R^{n_a \times n_b}
\label{SO_lie_deriv}
\end{align}

\begin{figure}
 \centering
 \includegraphics[width=0.5 \columnwidth]{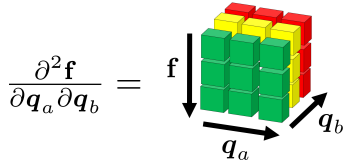}
\caption{Storage of elements of different variables in the 3D tensor}
 \label{f_tensor_block}
 \end{figure}

For a vector-valued function $\vf : \Qcal \rightarrow \R^m$, the derivative $\frac{\partial^2 \vf}{\partial \q_a \partial \q_b} \in \R^{m \times n_a \times n_b} $ is a tensor, which results from the component-wise second-order Lie derivatives. Figure~\ref{f_tensor_block} shows the arrangement of the elements of each of the variables in the 3D tensor. The elements of the function $\vf$ vary along the rows, while derivatives along the free modes for $\q_{a}$ and $\q_{b}$ vary along the columns and the pages of the tensor. In this paper, dimensions 1, 2 and 3 also refer to the dimensions along the rows, columns, and pages.

\begin{figure}[t]
 \centering
 \includegraphics[width=0.8 \columnwidth]{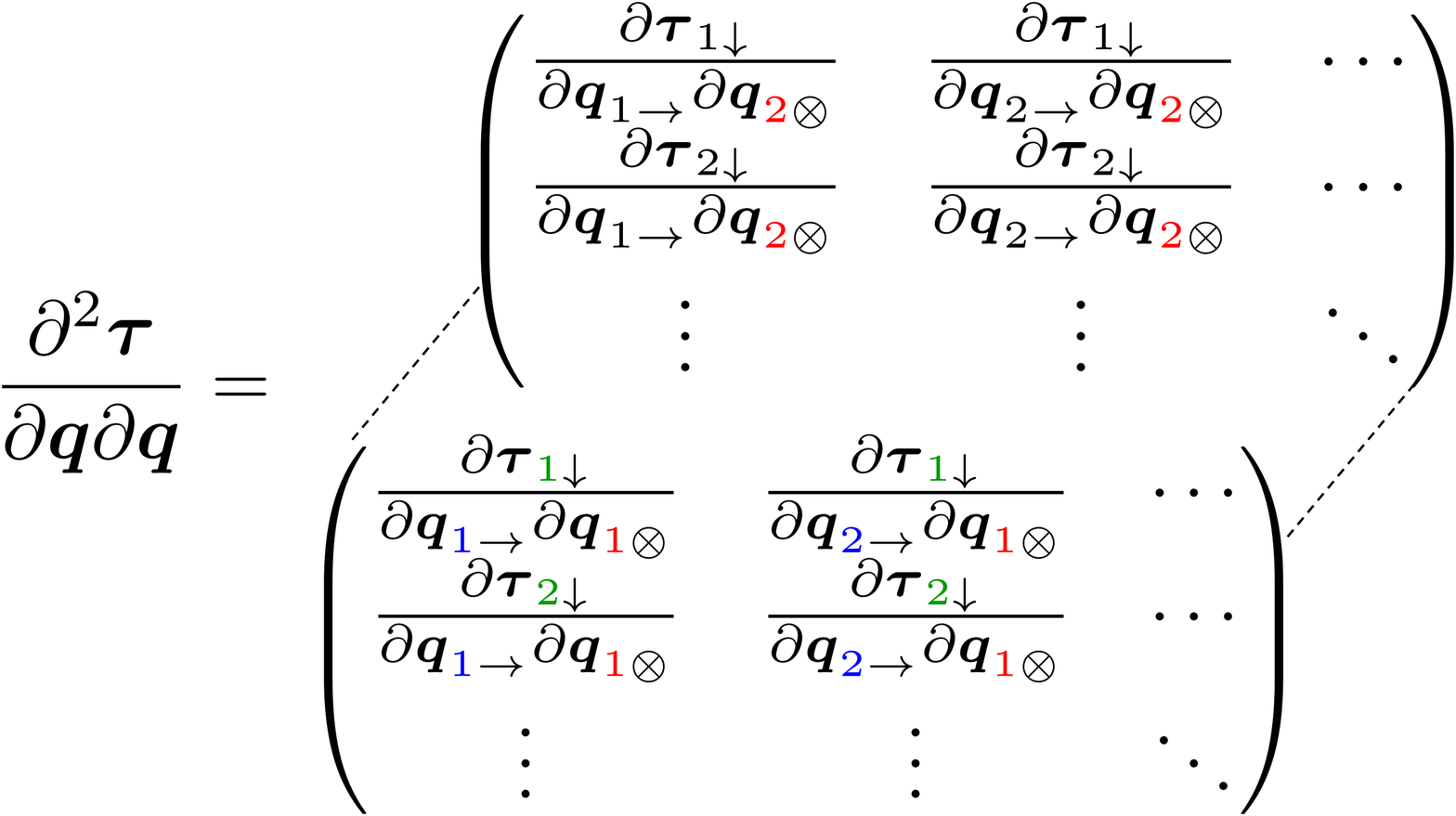}
\caption{An example of the ID SO tensor storage order. The down-arrow ($\downarrow$) shows $\taubar_{1},\taubar_{2}$ varies along the rows. The right-arrow($\rightarrow$)  shows that the $\q_{1},\q_{2}$ vary along the columns, while the otimes ($\otimes$) symbol shows that the elements of $\q_{1},\q_{2}$ vary along the pages of the tensor.}
 \label{tau_tensor_expand}
 \end{figure}

Returning to our dynamics example, we now consider the tensor block $\frac{\partial^2 \taubar_{i}}{\partial \q_{j} \partial \q_{k}}$. When joints $j$ and $k$ are single-DoF joints, each $\frac{\partial^2 \taubar_{i}}{\partial \q_{j} \partial \q_{k}}$ represents a conventional SO derivative w.r.t joint variables. In this case, the order of $\q_{j}$ and $\q_{k}$ doesn't matter. This operation becomes more nuanced when considering derivatives w.r.t configuration for a multi-DoF joint. 
When the Lie derivatives along the free modes of a joint do not commute, the order of $\q_{j}$ and $\q_{k}$ matters, and the tensor block $\frac{\partial^2 \taubar_{i} }{\partial \q_{j} \partial \q_{k}}$ can lack the usual symmetry properties. Such a case occurs, for example, with a spherical joint, since rotations do not commute. The tensor block for this case is similar to that shown in Fig.~\ref{f_tensor_block}, with $\taubar$ instead of $\vf$. An expanded view of the full ID SO tensor is shown in Fig.~\ref{tau_tensor_expand} with different tensor blocks embedded in it. Elements of $\taubar$ and $\q$ vary along the rows, columns, and pages, as depicted by the arrows and the $\otimes$ symbol.


In the following sections, we extend the partial derivatives of ID and FD to second order using \eqref{dtau_dq}-\eqref{dtau_dqd}. Given the tensorial nature of the second-order derivatives, we first consider an extension to spatial vector algebra that enables working with the derivative tensors.

\section{Tensorial Spatial Vector Algebra}
\label{sec:SVA_tensor}
\begin{figure}
\center
\includegraphics[width=0.8 \columnwidth]{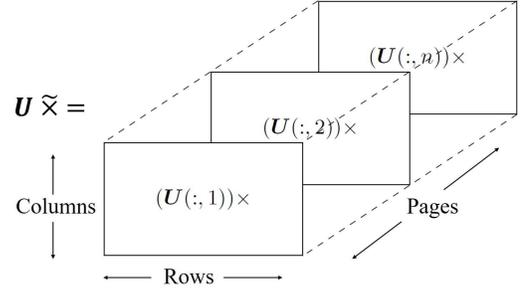}
\caption{$\timesM$ operates on each column of a $\U \in M^{6\times n}$ spatial matrix to create a third-order tensor. Each rectangular box is a 2D matrix} \label{fig:crossM}
\vspace{-5px}
\end{figure}

The motion space $M^{6}$~\cite{Featherstone08} is extended to a space of spatial-motion matrices $M^{6\times n}$, where each column of such a matrix is a usual spatial motion vector.
For any $\U \in M^{6 \times n}$, a new spatial cross-product operator $ \timesM$ is considered and defined by applying the usual spatial cross-product operator ($\times$) to each column of $\U$. The result is a third-order tensor (Fig.~\ref{fig:crossM}) in $\R^{6 \times 6 \times n}$ where each $6 \times 6$ matrix in the 1-2 dimension is the original spatial cross-product operator on a column of $\U$. 

Given two spatial motion matrices, $\U \in M^{6 \times n_u}$ and $\V \in M^{6 \times n_v}$ , we can now define a cross-product operation between them as $(\U \timesM) \V \in M^{6 \times n_v \times n_u}$ via a tensor-matrix product. Such an operation, denoted as $\mathcal{Z} = \mathcal{A} \B$ is defined as:
\begin{equation}
    \mathcal{Z}_{ijk} = \sum_{\ell}\mathcal{A}_{i \ell k}\\B_{\ell j}
\end{equation}
for any tensor $\mathcal{A}$ and suitably sized matrix $\B$. Thus, the $k$-th page, $j$-th column of $\U \timesM \V$ gives the cross product of the $k$-th column of $\U$ with the $j$-th column of $\V$.

In a similar manner, consider a spatial force matrix $\F\in F^{6 \times n_f}$. Defining $(\V\timesfM)$ in an analogous manner to in Fig.~\ref{fig:crossM} allows taking a cross-product-like operation $\V\timesfM \F$. Again, analogously, we consider a third operator $(\F\crffM)$ that provides  $\V\timesfM \F = (\F\crffM \V)\Rten$, where $(\Rten)$ is a tensor transpose swapping columns and pages, as precisely specified below. Here each page of $\F \crffM$ gives the $(\crff)$ operator on a column of the spatial force matrix $\F$. In each case, the tilde indicates the spatial-matrix extension of the usual spatial-vector cross products. The body-Coriolis matrix in \eqref{bk_term_defn} is now extended with a spatial matrix argument as: 
\begin{equation}
    \Bten{\I}{\V}  = \frac{1}{2} \big[ \big(\V\timesfM \big)\I - \I \big(\V \timesM   \big) + \big(\I \V \big) \crffM    \big]
    \label{bl_psijdot_term_defn}
\end{equation}
The result is a $6\times 6 \times n$ tensor, where each page is a body-Coriolis matrix associated with a single column of $\V$.

For later use, the product of a matrix $\B \in \R^{n_{1} \times n_{2}}$, and a tensor $\mathcal{A}\in \R^{n_{2} \times n_{3} \times n_{4}}$, likewise results in another tensor, denoted as $\mathcal{Y} = \B\mathcal{A}$, and defined as:

\begin{equation}
    \mathcal{Y}_{ijk} = \sum_{\ell}\B_{i \ell}\mathcal{A}_{\ell j k}
    \label{mat_ten_eqn}
\end{equation}

\begin{figure}
  \centering
  \subfloat[$\mathrm{(\Tten)}$ 1-2 Transpose]{\includegraphics[width=0.2\textwidth]{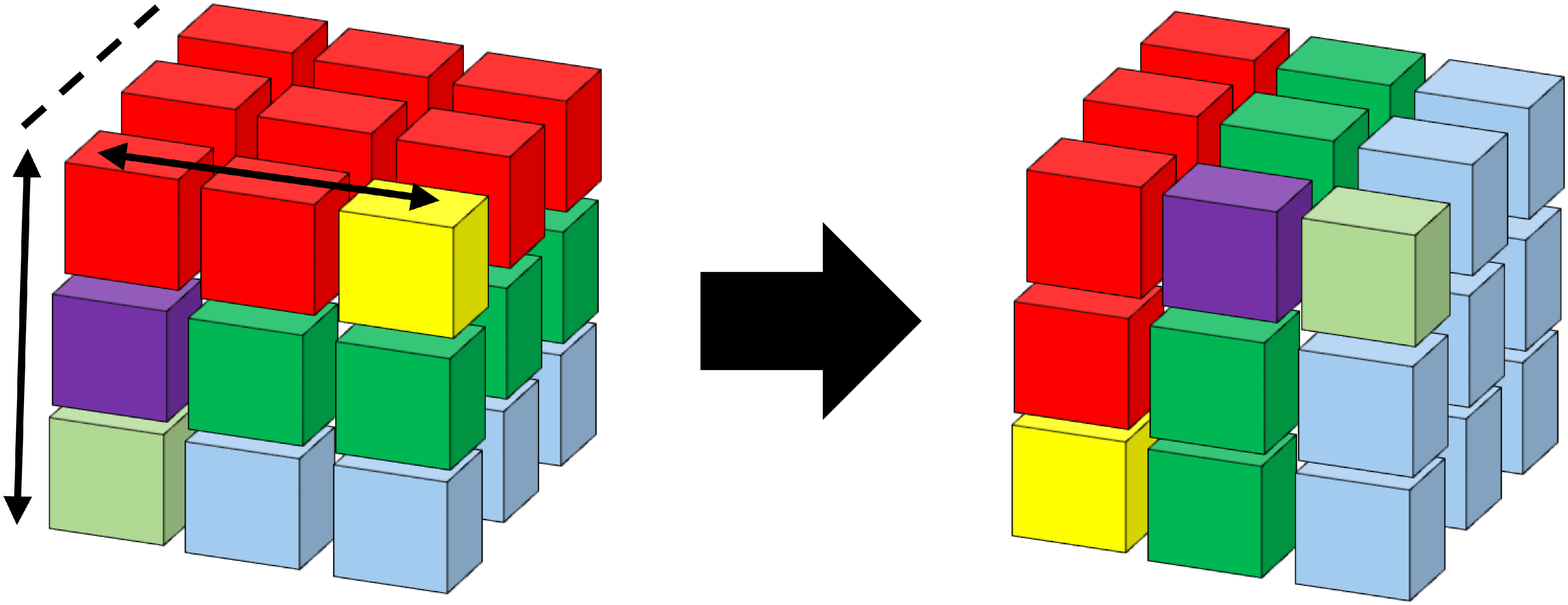}\label{rotation_operT}}
 \hspace{0.5cm}
  \subfloat[$\mathrm{(\Rten)}$ 2-3 Transpose]{\includegraphics[width=0.2\textwidth]{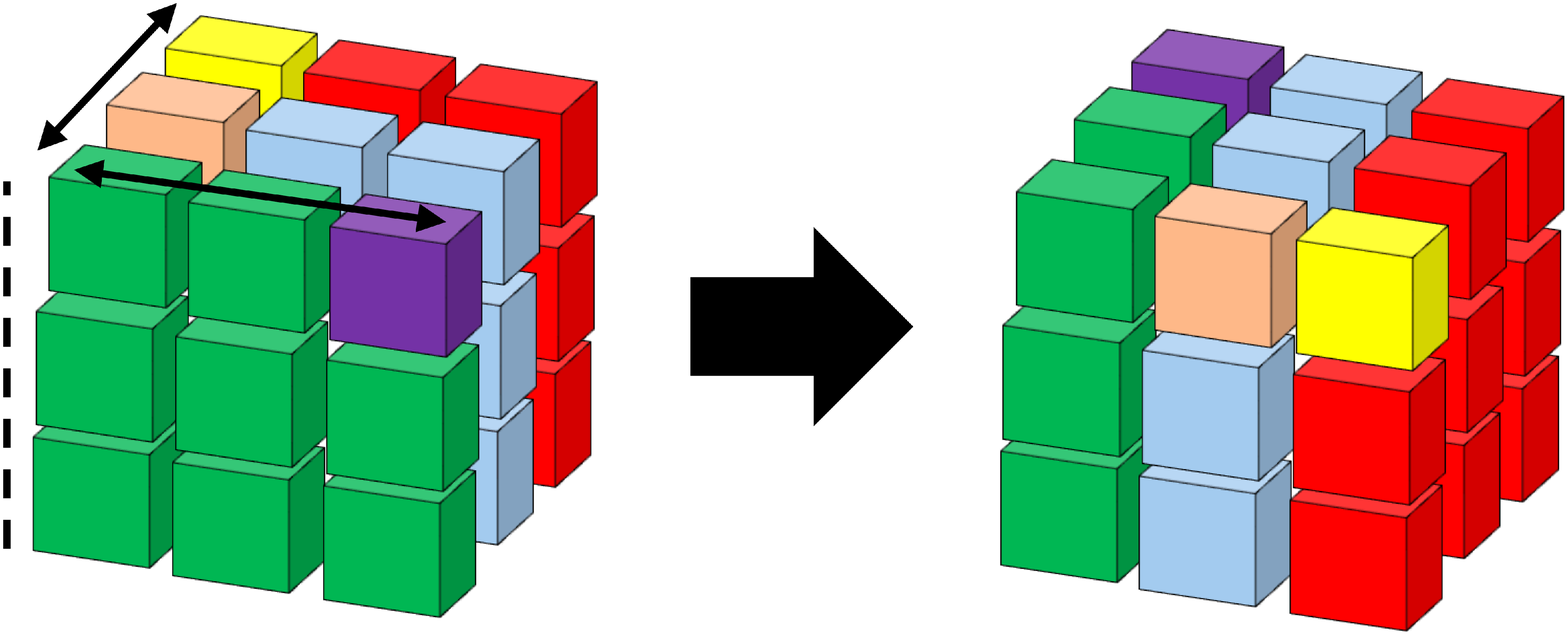}\label{rotation_operR}}
  \caption{Two types of tensor transposes}
\end{figure}

Two types of tensor transposes are defined for this paper:
\begin{enumerate}
    \item  $\mathcal{A}\Tten$: Transpose along the 1-2 dimension. This operation can also be understood as the usual matrix transpose of each matrix (e.g., in Fig.~\ref{rotation_operT}) moving along pages of the tensor. If $ \mathcal{A}\Tten = \mathcal{B}$, then $\mathcal{A}_{i,j,k} = \mathcal{B}_{j,i,k}$.\\[-1.7ex]
    \item  $\mathcal{A}\Rten$: Transpose of elements along the 2-3 dimension (Fig.~\ref{rotation_operR}). If $ \mathcal{A}\Rten = \mathcal{B}$, then $\mathcal{A}_{i,j,k} = \mathcal{B}_{i,k,j}$.
\end{enumerate}
Another transpose ($\RTten$) is a combination of  $(\Rten)$ followed by $(\Tten)$. For example, if $\mathcal{A}\RTten = \mathcal{B}$, then $\mathcal{A}_{i,j,k} = \mathcal{B}_{k,i,j}$.

Properties of the operators are given in Table~\ref{tab:SVM_props}. These properties naturally extend spatial vector properties~\cite[Table 2.1]{Featherstone08}, but with the added book-keeping required from using tensors. For example, the spatial force/vector cross-product operator $\timesf$ satisfies $\v\timesf = -\v\times \T$. Property \ref{m1} provides the matrix analogy for the spatial matrix operator $\timesfM$. Property \ref{m8} is the spatial matrix analogy of the spatial vector property $\u \times \v = -\v \times \u$, where $\u$ and $\v$ are spatial vectors. The verification of these identities is conceptually straightforward, but practically tedious. As such, we provide unit tests to verify them for the reader in lieu of derivations~\cite{matlab_prop_unit}.

\begin{table}[t]
\center 
\fbox{
\begin{minipage}{.88\columnwidth}
{\small
\begin{enumerate}[label=M{{\arabic*}})]
 \setlength\itemsep{.15em}
    \item \label{m1} $\U \timesfM = -(\U \timesM)\Tten$ 
    \item \label{m2} $-\Uv\T (\Uu \timesfM) = (\Uu \timesM \Uv)\Tten $
    \item \label{m3} $-\Uv\T (\Uu \timesfM) \Ff= (\Uu \timesM \Uv)\Tten \Ff$
    \item \label{m4} $ (\U \timesM \v)\Rten = - \v \times \U $ 
    \item \label{m5} $\U \timesfM \Ff = ( \Ff \crffM \U)\Rten$ 
    \item \label{m6} $\Ff \crffM \U = (\U \timesfM \Ff)\Rten $ 
    \item \label{m7} $(\lambda \U) \timesM = \lambda(\U \timesM)$
    \item \label{m8} $\Uu \timesM \Uv = - (\Uv \timesM \Uu)\Rten $ 
    \item \label{m9} $(\v \times \U) \timesM = \v \times \U \timesM - \U \timesM \v \times$ 
    \item \label{m10} $(\v \times \U) \timesfM = \v \times^* \U \timesfM - \U \timesfM \v \times^*$
    \item \label{m11} $((\U \timesfM \f)\Rten) \crffM = \U \timesfM \f \crff - \f \crff \U \timesM$
    \item \label{m12} $(\U \timesfM \Ff)\Tten = -\Ff\T (\U \timesM)$
    \item \label{m13} $\Uv\T (\Uu \timesfM \Ff) = (\Uv \timesM \Uu)\RTten \Ff=  (\Ff\T (\Uv \timesM \Uu)\Rten)\Tten $
    \item \label{m14} $ \v \times^* \Ff = (\Ff \crffM \v)\Rten$
    \item \label{m15} $ \f \crff \U = (\U \timesfM \f)\Rten$
     \item \label{m16} $\Uv\T (\Uu \timesfM \Ff)\Rten = \big[(\Uv \timesM \Uu)\RTten \Ff\big]\Rten$
          \item \label{m17} $\Uv\T (\Uu \timesfM \Ff)\Rten =-[\Uu\T (\Uv \timesfM \Ff)\Rten]\Tten$
      \item \label{m18} $\I  (\Uu \timesfM \Ff)\Rten = [ \I (\Uu \timesfM \Ff)]\Rten$ 
        \item \label{m19} $\I  (\Uu \timesM \Uv)\Rten = [ \I (\Uu \timesM \Uv)]\Rten$ 
    \item \label{m20} $(\A\mathcal{Y})\Tten = \mathcal{Y}\Tten \A\T$
            \item \label{m21} $\Ff\T(\U \timesM \V) = -[\V\T(\Ff\crffM\U)\Rten]\Tten$
       \item \label{m22} $\Bmat{\I}{\v}\T\w = -\Bmat{\I}{\w}\T\v$
        \item \label{m23}  $\Bmat{\I}{\v}\w= \Bmat{\I}{\w}\v-\I(\v\times \w)$
   \item \label{m24} $\u\T[\Bmat{\I}{\v}\w] = -\v\T[\Bmat{\I}{\u}\w]$
          \item \label{m25} $\Bten{\I}{\V}\Tten\W = -[\Bten{\I}{\W}\Tten\V]\Rten$
   \item \label{m26} $\Bten{\I}{\V}\W = [\Bten{\I}{\W}\V]\Rten - \I(\V\timesM)\W$
          \item \label{m27} $\U\T[\Bten{\I}{\V}\W]\Rten = -\Big[\V\T[\Bten{\I}{\U}\W]\Rten\Big]\Tten$

\end{enumerate}
}
\end{minipage}
}
\caption{Spatial Matrix Algebra Properties: $\u, \v, \w \in M^{6}$,$\f \in F^{6}$, $\U \in M^{6 \times n}$, $\Uv \in M^{6 \times l}$, $\W \in M^{6 \times p}$, $\Ff \in F^{6 \times m} $, $\A \in \R^{n_{1} \times n_{2}}$, $\I, \B \in \R^{6 \times 6}$, $\mathcal{Y} \in \R^{n_{2} \times n_{3} \times n_{4}}$, $\mathcal{B} \in \R^{6\times 6 \times n_{5}}$.}
\label{tab:SVM_props}
\end{table}


\section{SO Derivatives of Inverse Dynamics: Theory}
\label{sec:ID_SO_theory}

\subsection{Preliminaries}
\label{prem_sec}

In this section, we will consider the second-order partial derivatives of ${\rm ID}({\rm model}, \q,\qd,\qdd)$ \eqref{inv_dyn_model} which, again, we view as a function ${\rm ID} : \Qcal\times \R^n \times \R^n \rightarrow \R^n$. We denote the partial derivatives of this function as $\frac{\partial^2 \taubar}{\partial \u \partial \w}$ where $\u$ or $\w$ represents $\q$, $\qd$, or $\qdd$.
Many of these second-order partials are zero, limiting the cases to be considered. From \eqref{inv_dyn}, the first-order partial derivative of $\taubar{}$ w.r.t $\qdd$ is ${\partial \taubar}/{\partial \qdd} = \M(\q)$. Taking subsequent partial derivatives results in ${\partial ^2 \taubar}/{\partial \qdd \partial \qdd}=\mathbf{0}$ and  ${\partial^2 \taubar}/{\partial \qdd \partial \qd}=\mathbf{0}$. However, the cross-derivative w.r.t $\qdd$ and $\q$ is non-trivial and equals ${\partial \M(\q)}/{\partial \q}$. Garofalo et al.~\cite{garofalo2013closed} present formulas for the partial derivative of $\M(\q)$ w.r.t $\q$ for multi-DoF joints. In this work, we re-derive that result using newly developed spatial matrix operators and contribute new analytical SO partial derivatives of ID w.r.t $\q$ and $\qd$.


 To obtain the partial derivatives of spatial quantities embedded in \eqref{dtau_dq}-\eqref{dtau_dqd}, a number of derivative identities (\appref{mdof_iden}) are derived assuming the quantities in the ground frame. These are an extension to ones defined in Ref.~\cite{singh2022efficient}, but use the newly developed spatial matrix operators from Sec.~\ref{sec:SVA_tensor}. Step by step derivation is provided in~\cite{arxivss_SO} for the interested reader, with unit tests available online~\cite{matlab_iden_unit}.
 
 For example, identity \ref{dphii} is an extension of identity J1 in Ref.~\cite{singh2022efficient}. The identity J1 \eqref{J1_eqn} gives the directional derivative of the joint motion sub-space matrix $\Phibar_{i}$ w.r.t the $p^{th}$ DoF of a previous joint $j$ in the connectivity tree:
\begin{equation}
\frac{\partial \Phibar_{i}}{\partial \q_{j,p}}\triangleq \Lie{X_{j,p}} \Phibar_i =  \phibar_{j,p}   \times \Phibar_{i} ~~~ (j \preceq i)
\label{J1_eqn}
\end{equation}
%
On the other hand, \ref{dphii} uses the $\timesM$ operator to extend it to the partial derivative of $\Phibar_{i}$ w.r.t all free modes for $\q_{j}$ to give the tensor $\frac{\partial \Phibar_{i}}{\partial \q_{j}}$ as:
\begin{equation}
\frac{\partial \Phibar_{i}}{\partial \q_{j}}=  \Phibar_{j}   \timesM \Phibar_{i}   ~~~ (j \preceq i) 
\end{equation}
Individual partial derivatives of these quantities allow us to use the plug-and-play approach to simplify the algebra needed for SO partial derivatives of ID.

To calculate the SO partial derivatives, we use \eqref{dtau_dq} and \eqref{dtau_dqd}, and take subsequent partial derivatives of the terms $\frac{\partial \taubar_{i}}{\partial \q_{j}}$, $\frac{\partial \taubar_{j}}{\partial \q_{i}}$,$\frac{\partial \taubar_{i}}{\partial \qd_{j}}$, and  $\frac{\partial \taubar_{j}}{\partial \qd_{i}}$ w.r.t joint configuration ($\q$), and joint velocity ($\qd$) for joint $k$. 
We consider three cases, by changing the order of index $k$ as:
\[
\textbf{Case~A:}~ k \preceq j \preceq i ~~~
\textbf{Case~B:}~  j \prec k \preceq i ~~~
\textbf{Case~C:}~ j \preceq i \prec k
\]
The sections below outline the approach for calculating the SO partial derivatives. Only some cases are shown to illustrate the main idea of the approach, with a detailed summary of all the cases in \appref{summary_sec}, and full step-by-step derivations in Ref.~\cite{arxivss_SO}. 
Given the extent of the full derivation, we again provide unit tests~\cite{matlab_SO_unit} to verify each of the results empirically. The reader not interested in the derivation strategy may wish to skip to Sec.~\ref{sec:IDSVA_SO} to review the final algorithm. 
\subsection{Second-order partial derivatives w.r.t $\q$}
\label{sec:ID_SO_q}
For SO partials of ID w.r.t $\q$, we take the partial derivatives of \eqref{dtau_dq_1}, and~\eqref{dtau_dq_2} w.r.t $\q_{k}$ for cases A, B, and C mentioned previously. However, the symmetry in the Hessian allow us to re-use three of those six cases.  For any of the cases, the partial derivatives of \eqref{dtau_dq_1} and~\eqref{dtau_dq_2} can be taken, as long as the accompanying conditions on the equations are met. For example, when using Case C ($j \preceq i \prec k$) on \eqref{dtau_dq_2},  the accompanying condition $j \prec i$ modifies C into a stricter case ($j \prec i \prec k$). A derivation for Case C is shown here as an example. We take the partial derivative of \eqref{dtau_dq_1} w.r.t $\q_{k}$. Applying the product rule, and using the identities \ref{psii_dot}, \ref{Psiddoti}, and \ref{PhiT} (App.~\ref{mdof_iden}) as:
\begin{equation}
  \frac{\partial^2 \taubar_{i}}{\partial \q_{j} \partial \q_{k}} = 2\Phibar_{i} \T \left(  \frac{\partial \B_{i}^{C}}{\partial \q_{k}}\Psibardot{}_{j} \right) +  \Phibar_{i} \T  \left(\frac{\partial \I_{i}^{C}}{\partial \q_{k}}\Psibarddot{}_{j} \right)
\end{equation}
Using the identities \ref{Iic} and \ref{Bic} then  gives:
\begin{align}
  &\frac{\partial^2 \taubar_{i}}{\partial \q_{j} \partial \q_{k}} = 2\Phibar_{i} \T \Big(   \Bten{\I_{k}^{C}}{\Psibardot_{k}}+\Phibar_{k} \timesfM \B_{k}^{C}-  \B_{k}^{C}(\Phibar_{k} \timesM) \Big)\Psibardot_{j} \nonumber\\ &
  ~~~~~+ \Phibar_{i} \T  \Big( \Phibar_{k} \timesfM \I_{k}^{C}- \I_{k}^{C}(\Phibar_{k} \timesM)\Big) \Psibarddot_{j}, \hspace{0.5cm} (j \preceq i \prec k) 
    \label{SO_q_case2_eqn1}
  \end{align}
Expressions for other cases are in \appref{summary_sec} with full derivation in Ref.~\cite[Sec.~IV]{arxivss_SO}.
 
\subsection{Cross Second Order Partial derivatives w.r.t $\qdd$ and $\q$}
\label{sec:ID_MSO_q_qdd}

As explained before, the cross-SO partial derivatives of ID w.r.t $\qdd$ and $\q$ results in $\frac{\partial \M}{\partial \q}$. The lower-triangle of the mass matrix $\M(\q)$ for the case $j \preceq i$ is given as~\cite{Featherstone08}:
\begin{equation}
    \M_{ji} = \Phibar_{j}\T \IC{i} \Phibar_{i}
 \label{M_eqn}
\end{equation}
Since $\M(\q)$ is symmetric~\cite{Featherstone08}, $\M_{ij} = \M_{ji}\T$. We apply the three cases A, B, and C discussed before. As an example, for Case B ($j \prec k \preceq i$), we take the partial derivative of $\M_{ji}$ w.r.t $\q_{k}$ and use the product rule, along with identity \ref{PhiT} as:
\begin{equation}
    \frac{\partial \M_{ji}}{\partial \q_{k}} =  \Phibar_{j}\T \left( \frac{\partial \IC{i} }{\partial \q_{k}}\Phibar_{i} +\IC{i}\frac{\partial \Phibar_{i}}{\partial \q_{k}} \right)
\end{equation}
Using with identities \ref{dphii} and \ref{Iic}, and canceling terms gives:
\begin{equation}
        \frac{\partial \M_{ji}}{\partial \q_{k}} =  \Phibar_{j}\T ( \Phibar_{k} \timesfM \I_{i}^{C})\Phibar_{i}, \hspace{0.5cm} (j \prec k \preceq i)
        \label{M_FO_eqn1}
\end{equation}
 Expressions for other cases are listed in  \appref{summary_sec}, with details of derivation at Ref.~\cite[Sec. VII]{arxivss_SO}.
\parspace
\subsection{Second-order partial derivatives w.r.t. $\qd$}
\label{sec:ID_SO_qd}

For SO partial derivatives w.r.t $\qd$, we take the partial derivatives of \eqref{dtau_dqd_1} and~\eqref{dtau_dqd_2} w.r.t $\qd_{k}$. Here, the Case A is split into two cases: 1) $k \prec j \preceq i$, and 2) $ k=j \preceq i$. This split arises due to the condition for using identity \ref{Psidot_qdj}, and makes the associated algebra easier to follow. A similar split also occurs for Case B into 1) $ j \prec k \prec i$, and 2) $ j \prec k = i$. Since now the total number of cases is five, this results in a total of eight expressions. Six of these expressions result from the three main cases A, B and C, while the other two are from the split explained above. Hessian symmetry allows us to re-use three of these eight expressions. 

Here, the details of one of the cases $k \prec j \preceq i$ are given. We take the partial derivative of \eqref{dtau_dqd_1} w.r.t $\qd_{k}$ as:
\begin{equation}
    \frac{\partial^2 \taubar_{i}}{\partial \qd_{j} \partial \qd_{k}} = \Phibar_{i}\T \Bigg[2 \frac{\partial \B_{i}^{C}}{\partial \qd_{k}} \Phibar_{j} +  \I_{i}^{C} \Bigg(   \frac{\partial \Psibardot_{j}}{\partial \qd_{k}}  + \frac{\partial \Phibardot_{j} }{\partial \qd_{k}} \Bigg) \Bigg]
        \label{SO_qd_case1A_eqn1}
\end{equation}
Using the identities \ref{Phidot_qdj}-\ref{Bic_qdj}, and simplifying:
\begin{equation}
    \frac{\partial^2 \taubar_{i}}{\partial \qd_{j} \partial \qd_{k}} = 2 \Phibar_{i}\T \Big[ \Bten{\I_{i}^{C}}{\Phibar_{k}} \Phibar_{j} +  \I_{i}^{C} \big(   \Phibar_{k} \timesM \Phibar_{j}   \big) \Big]
    \label{SO_qd_case1A_eqn2}
\end{equation}
 In \eqref{SO_qd_case1A_eqn2}, the term $\Bten{\I_{i}^{C}}{\Phibar_{k}}$ \eqref{bl_psijdot_term_defn}
has terms with indices $i$ and $k$ inter-mingled, which makes it harder to implement. 
Hence, we re-write the expression in a form where the terms of a particular index are grouped together.

Using \ref{m26}
\begin{align}
    &\frac{\partial^2 \taubar_{i}}{\partial \qd_{j} \partial \qd_{k}} = 2 \Phibar_{i}\T \Big[ (\Bten{\I_{i}^{C}}{\Phibar_{j}}  \Phibar_{k})\Rten -\I_{i}^{C} (   \Phibar_{k} \timesM \Phibar_{j}   )+ \\
    &~~~~~~~~~~~~~~~~~~~~~~~\I_{i}^{C} (   \Phibar_{k} \timesM \Phibar_{j}   ) \Big]
\end{align}
Cancelling terms and using \ref{m27} results in:
\begin{equation}
    \frac{\partial^2 \taubar_{i}}{\partial \qd_{j} \partial \qd_{k}} =-\big[\Phibar_{j}\T (2\Bten{\I_{i}^{C}}{\Phibar_{i}} \Phibar_{k})\Rten \big]\Tten ,   (k \prec j \preceq i)     \label{SO_qd_case1A_eqn7}   
\end{equation}
This form now allows the computation of the $i$-index term $\Bten{\I_{i}^{C}}{\Phibar_{i}}$ at once, followed by the operation of $\Phibar_{j}$ and $\Phibar_{k}$ from the left and right. Similar algebraic manipulation using the identities in Table~\ref{tab:SVM_props} is performed extensively throughout to simplify the expressions for an efficient implementation. A list of expressions for the rest of the cases is given in  \appref{summary_sec}, with full derivation in Ref.~\cite[Sec. V]{arxivss_SO}.

\subsection{Cross Second order partial derivatives w.r.t $\q$,$\qd$}
\label{sec:ID_SO_q_qd}

For cross-SO partial derivatives of ID, we take the partial derivative of Eqs.~\ref{dtau_dqd_1}-\ref{dtau_dqd_2} w.r.t $\q_{k}$ to get $\frac{\partial^2 \taubar}{\partial \qd \partial \q}$. The three cases A, B, and C (Sec.~\ref{prem_sec}) for \eqref{dtau_dqd_1}-\eqref{dtau_dqd_2} result in six expressions, which are then also used for the symmetric term $\frac{\partial^2 \taubar}{\partial \q \partial \qd}$ as:
\begin{equation}
    \frac{\partial^2 \taubar}{\partial \q \partial \qd} = \left[\frac{\partial^2 \taubar}{\partial \qd \partial \q} \right]{\!\vphantom{\Big)}}\Rten
\end{equation}
Here, we solve all the three cases A, B and C for both \eqref{dtau_dqd_1} and \eqref{dtau_dqd_2}.
Pertaining to Case A ($k \preceq j \preceq i$), since $j \preceq i$, \eqref{dtau_dqd_1} can be safely used to get $ \frac{\partial^2 \taubar_{i}}{ \partial \qd_{j}\partial \q_{k}}$. However, the $j \ne i$ requirement on \eqref{dtau_dqd_2} constrains the condition in Case A to $k \preceq j \prec i$. Similarly, for Case C ($j \preceq i \prec k$), taking partial derivative of \eqref{dtau_dqd_2} results in a stricter case $j \prec i \prec k$. 
Appendix \ref{summary_sec} lists the six expressions with full derivation in Ref.~\cite[Sec. VI]{arxivss_SO}.
\parspace
\subsection{Efficient Implementation Techniques}
For efficient implementation of the algorithm, all the cases are converted to an index order of $k \preceq j \preceq i$. This process is explained with the help of following two examples.

{\em Example 1:} In \eqref{SO_q_case2_eqn1}, we first we switch the indices $k$ and $j$, followed by $j$ and $i$ to get $\frac{\partial^2 \taubar_{j}}{\partial \q_{k} \partial \q_{i}}$ as
%
\begin{align}
  &\frac{\partial^2 \taubar_{j}}{\partial \q_{k} \partial \q_{i}} = 2\Phibar_{j} \T \Big(   \Bten{\I_{i}^{C}}{\Psibardot_{i}}+\Phibar_{i} \timesfM \B_{i}^{C}-   \B_{i}^{C}(\Phibar_{i} \timesM) \Big)\Psibardot_{k} \nonumber\\
  &~~~~~~~ + 
  \Phibar_{j} \T  \Big( \Phibar_{i} \timesfM \I_{i}^{C}-  \I_{i}^{C}(\Phibar_{i} \timesM)\Big) \Psibarddot_{k}, (k \preceq j \prec i) \label{tau_SO_q_1C}
\end{align}
For the term  $\frac{\partial^2 \taubar_{j}}{\partial \q_{i} \partial \q_{k}}$, when $k \ne i$, the symmetry property of Hessian blocks can be exploited:
\begin{equation}
    \frac{\partial^2 \taubar_{j}}{\partial \q_{i} \partial \q_{k}} =  \left[  \frac{\partial^2 \taubar_{j}}{\partial \q_{k} \partial \q_{i}} \right]{\!\vphantom{\Big)}}\Rten, (k \preceq j \prec i)
      \label{tau_SO_q_2C}
\end{equation}
The 2-3 tensor transpose in \eqref{tau_SO_q_2C} occurs due to symmetry along the 2\textsuperscript{nd} and 3\textsuperscript{rd}  dimensions.

{\em Example 2:}
\noindent In \eqref{M_FO_eqn1}, switching indices $k$ and $j$ leads to the index order $k \prec j \preceq i$. Using identity \ref{m5} leads to:
\begin{equation}
       \frac{\partial \M_{ki}}{\partial \q_{j}} =  \Phibar_{k}\T ((\I_{i}^{C}\Phibar_{i}) \crffM \Phibar_{j})\Rten   , (k \prec j \preceq i)
     \label{M_FO_case_1B}
\end{equation}

\noindent Symmetry of $\M(\q)$ gives us $ \frac{\partial \M_{ik}}{\partial \q_{j}}$ as:
\begin{equation}
\small
      \frac{\partial \M_{ik}}{\partial \q_{j}} = \left[\frac{\partial \M_{ki}}{\partial \q_{j}}  \right]\Tten 
     \label{M_FO_case_2B}
\end{equation}

\noindent In this case, since the symmetry is along the 1\textsuperscript{st} and the 2\textsuperscript{nd} dimension of $\frac{\partial \M_{ki}}{\partial \q_{j}} $, the tensor 1-2 transpose takes place.

The expressions for SO partials of ID (\appref{summary_sec}) are first reduced to matrix and vector form to avoid tensor operations. This refactoring is due, in part, to a lack of stable tensor support in the C++ Eigen library that is often used in robotics dynamics libraries. This reduction is achieved by considering the expressions for single DoF of joints $i$, $j$, and $k$, one at a time. We explain this process with the help of two examples. 

{\em Example 1}: Considering the case $k \prec j \preceq i$ from \eqref{SO_qd_case1A_eqn7}
\[
\small
\frac{\partial^2 \taubar_{i}}{\partial \qd_{j} \partial \qd_{k}} =-\big[\Phibar_{j}\T (2\Bten{\I_{i}^{C}}{\Phibar_{i}} \Phibar_{k})\Rten \big]\Tten
\]
 is studied for the $p^{th}$, $t^{th}$, and $r^{th}$ DoFs of the joints $i$, $j$ and $k$, respectively. The tensor term $\Bten{\I_{i}^{C}}{\Phibar_{i}}$ (defined by \eqref{bl_psijdot_term_defn}) reduces to a matrix $\Bmat{\I_{i}^{C}}{\phibar_{i,p}}$, where $\phibar_{i,p}$ is $p$-th column of $\Phibar_i$. This term represents the value $\B_i^C$ would take if all bodies in the subtree at $i$ moved with velocity $\phibar_{i,p}$.
 The above reduction enables dropping the 3D tensor transpose $(\Rten)$. The products of $\Bmat{\I_{i}^{C}}{\phibar_{i,p}}$ with column vectors $\phibar_{j,t}$ and $\phibar_{k,r}$ then provides a scalar, resulting in dropping the transpose $(\Tten)$ from above: 

\begin{equation}
\small 
    \frac{\partial^2 \taubar_{i,p}}{\partial \qd_{j,t} \partial \qd_{k,r}} =- 2\phibar_{j,t}\T\big( \Bmat{\I_{i}^{C}}{\phibar_{i,p}}\phibar_{k,r}\big) 
    \label{eq:example1}
\end{equation}
%
{\em Example 2:}
The term $ \frac{\partial^2 \taubar_{i}}{\partial \qd_{k} \partial \qd_{j}} = \left[ \frac{\partial^2 \taubar_{i}}{\partial \qd_{j} \partial \qd_{k}} \right]\Rten$ is
evaluated for the $p^{th}$, $t^{th}$, and $r^{th}$ DoFs of the joints $i$, $j$ and $k$, respectively. In this case, the 2-3 tensor transpose ($\Rten$) drops out, since the resulting expression is a scalar.

\begin{equation}
\small
\frac{\partial^2 \taubar_{i,p}}{\partial \qd_{k,r} \partial \qd_{j,t}} =  \frac{\partial^2 \taubar_{i,p}}{\partial \qd_{j,t} \partial \qd_{k,r}}
\label{so_qd_eqn_2}
\end{equation}

\subsection{Final Algorithm}\label{sec:IDSVA_SO}
\begin{figure}
    \centering
      \includegraphics[width=0.5\columnwidth]{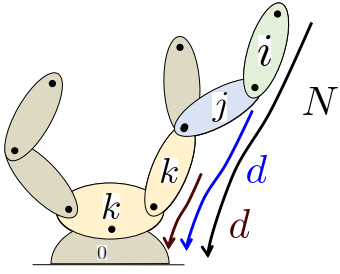}
    \caption{Backward-pass in IDSVA-SO (Alg.~\ref{alg:tau_SO_v5}) goes from leaves to root, $i$ cycles through all $N$ bodies, $j$ from $i$ to root, and $k$ from $j$ to root.}
 \label{algo_back_pass}
\end{figure}

Algorithm~\ref{alg:tau_SO_v5} (IDSVA-SO) is detailed in \appref{summary_sec} and returns all the SO partials from Sec.~\ref{sec:ID_SO_q}-\ref{sec:ID_SO_qd}.
The algorithm is implemented with all kinematic and dynamic quantities represented in the ground frame. 
The forward pass in Alg.~\ref{alg:tau_SO_v5} (Lines \ref{alg:forward_start}-\ref{alg:forward_end}) computes kinematic quantities like $\a_{i}$, $\Psibardot_{i}$, and $\Psibarddot_{i}$ and initializes dynamic quantities  $\BC{i}$, $\f{}_{i}^{C}$ for the entire tree. The factor $\frac{1}{2}$ is skipped from the definition of $\BC{i}$ to make expressions simpler, and necessary adjustments are made in the algorithm. The backward pass then cycles from leaves to root of the tree (see Fig.~\ref{algo_back_pass}) and consists of three main nested loops, each one over bodies $i$, $j$, and $k$. These three nested loops also consist of a nested loop for each DoF of joint $i$, $j$, and $k$. The indices $p$ (from 1 to $n_i$, Line \ref{alg:1b_start}), $t$ (from 1 to $n_j$, Line \ref{alg:2b_start}), and $r$ (from 1 to $n_k$, Line \ref{alg:3b_start}) cycle over all of the DoFs of joints $i$, $j$, and $k$ respectively. 
%
%
Some of the intermediate quantities in the algorithm are defined for a DoF of a joint. For example, in Line \ref{alg:1_define}, $\phibar_{p}$  
is simply the $p^{th}$ column of $\Phibar_{i}$. Similar quantities for joint $j$ and $k$ are defined in Line \ref{alg:2_define} and \ref{alg:3_define} respectively. 
The backward pass (Lines \ref{alg:1a_start}-\ref{alg:1a_end}) cycles  $n$ times in total (i.e., over all joints), with the nested loops (Lines \ref{alg:2b_start}-\ref{alg:2b_end} and \ref{alg:3b_start}-\ref{alg:3b_end}) each executing at most $6d$ times per cycle of its parent loop. Thus, the total computational complexity is $O(Nd^2)$.
%
%
While the algorithm is complex, an open source {\sc Matlab} version of it can be found at~\cite{matlabsource}, and is integrated with Featherstone's \texttt{spatial v2} library~\cite{Featherstone08}. 

Another version of the IDSVA SO algorithm was proposed in Ref.~\cite[v1]{singh2022closed}. Instead of the three nested loops for indices $i$, $j$, and $k$ in the backward pass, the SO expressions of that previous version were formulated for the entire sub-tree of the body $j$ ($\subtree(j)$, or $\subtreeb(j)$), resulting in a double-loop backward pass. This design was inspired by the approach used for the FO derivatives of ID~\cite{carpentier2019pinocchio,singh2022efficient}. In the SO case, this strategy has a negative effect on run-time performance for ID SO derivatives. For serial/branched chains with single-DoF revolute joints, the speed-up of the triple-loop approach over the double-loop is between $1.5-4 \times$. Due to this reason, the triple-loop version of the algorithm (as given in the appendix) is adopted for the rest of the run-time analysis herein. We also considered versions of the algorithm triple-loop algorithm that eliminated the iteration over individual DoFs of each joint. However, the algorithm as presented was found to give the fastest run times.

\section{SO Derivatives of Inverse Dynamics: Results}
\label{sec:ID_SO_results}

\subsection{Run-Time Analysis- Analytical vs Complex-Step, Chain-Rule \& Finite-Difference Methods}
\label{sec:ID_SO_alternative}
The run-time and accuracy performance of the analytical method presented is compared with several other methods. The Bi-complex method~\cite{lantoine2012using} is used with RNEA to get the ID SO derivatives (see App.~\ref{sec:bi_complex} for details). Two types of finite-difference methods based on the SO central-difference formulae are also compared. The Finite-Diff-1 method with complexity $\mathcal{O}(N^3)$ (App.~\ref{sec:finite_diff1}) uses central-difference formulae on RNEA, while Finite-Diff-2 with complexity $\mathcal{O}(N^2d)$ (App.~\ref{sec:finite_diff2}) uses a FO central-difference formula around the IDSVA~\cite{singh2022efficient} algorithm, which gives the FO partials of RNEA. 

\begin{figure}
 \centering
 \includegraphics[width=1.0 \columnwidth]{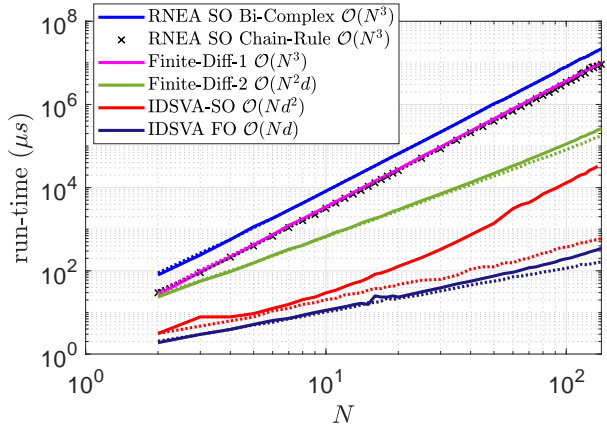}
 \caption{IDSVA-SO algorithm outperforms all the alternative methods for ID SO derivatives for serial chains (bold, cross), and binary trees (dotted). The run-time for IDSVA~\cite{singh2022efficient} (dark purple) is also provided for reference. All algorithms are implemented in Fortran 90, compiled using the IFORT Intel Fortran Compiler Classic in Windows 10.}
 \label{fort_ID_SO_v1}
 \end{figure}

\begin{figure}
 \centering
 \includegraphics[width=1.0 \columnwidth]{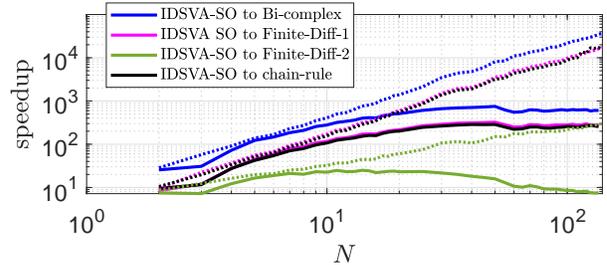}
 \caption{Speed-ups for the IDSVA-SO algorithm to alternative approaches. IDSVA-SO algorithm out-performs the Bi-Complex, Chain-Rule, and Finite difference methods in terms of run-times, for serial chains (bold lines), and binary trees (dotted lines).}
 \label{fort_ID_SO_speedup}
 \end{figure}

The run-time comparison of these methods for serial chains and binary trees, implemented in Fortran 90, with the IDSVA-SO algorithm is shown in Fig.~\ref{fort_ID_SO_v1}.
 To better understand the complexity of the algorithms, the run-time data is fitted in a Least-Squares sense to get the curve $\log(t) = A \log(N)+B$. The values of $A$ and $B$ are given in Tab.~\ref{tab:ID_SO_runtime_slope}, where the slope ($A$) gives the empirical polynomial order of the computational complexity of the algorithms, for serial-chains and binary trees. Each algorithm is called 10,000 times with randomized inputs for the joint configuration, velocity, and acceleration. The computations are performed on a 12th Gen Intel i5-12400 processor with a 2.5 GHz clock speed. The Finite-Diff-2 method outperforms Finite-Diff-1 by exploiting the faster FO analytical algorithm, resulting in a smaller run-time slope of $\approx 2.2$ (for binary trees) instead of 3 for the Finite-Diff-1, as shown in Tab.~\ref{tab:ID_SO_runtime_slope}. The bi-complex approach suffers in run-time due to the use of overloaded bi-complex numbers for intermediate variables in RNEA. The RNEA SO Chain-Rule is based on the forward SO chain-rule derivative of each line in the RNEA algorithm, with little to no simplifications. This is inspired by Lee et al.~\cite{lee2005newton}, who presented the SO Chain-Rule for Forward Dynamics derivatives. Although analytical, the Chain-Rule approach performs poorly due to a large number of matrix-matrix products resulting from the recursive chain-rule operations. As seen in Fig.~\ref{fort_ID_SO_speedup}, the IDSVA-SO algorithm outperforms the other methods discussed by more than an order of magnitude. Due to the branch-induced sparsity, the curves for serial chain and binary trees for the IDSVA-SO algorithm diverge. This is also clear from Tab.~\ref{tab:ID_SO_runtime_slope}, where the slope for the IDSVA-SO algorithm for binary trees results in a slope of $\approx 1.5$, as compared to a slope of $\approx 3.1$ for the serial chain case.

\begin{table}[]
\small
    \centering
    \begin{tabular}{c c  c c c}
    \hline
   Method &      \multicolumn{2}{c }{\bf Serial-Chains} &  \multicolumn{2}{c}{\bf Binary-Trees}  \\ 
       &  {\em $A$} & {\em $B$}   &  {\em $A$} & {\em $B$} \\ \hline 
         RNEA SO Bi-complex $\mathcal{O}(N^3)$ & 2.98 & 2.14  & 2.98 & 2.13 \\ \hline
         RNEA SO Chain-Rule $\mathcal{O}(N^3)$ & 3.05 & 0.93  & 3.17 & 0.47 \\ \hline
         Finite-Diff-1 $\mathcal{O}(N^3)$ & 3.02 & 1.16  & 3.03 & 1.13 \\ \hline
         Finite-Diff-2 $\mathcal{O}(N^2d)$ & 2.38 & 0.66  & \bf{2.15} & 1.38 \\ \hline
         IDSVA-SO $\mathcal{O}(Nd^2)$ & 3.08 & -4.71  & \bf{1.45} & -0.77 \\ \hline
         IDSVA-FO  $\mathcal{O}(Nd)$ & 1.76 & -2.06  & 1.03 & 0.01 \\ \hline
    \end{tabular}
    \caption{Coefficients of the equation  $\log(t) = A \log(N)+B$ for different ID SO derivatives approaches for serial chains and binary trees. For binary-trees, the Bi-complex, Chain-Rule, and Finite-Diff-1 demonstrate a complexity of $\mathcal{O}(N^3)$, but the effect of branch-induced sparsity can be seen in Finite-Diff-2 and IDSVA-SO algorithms}
    \label{tab:ID_SO_runtime_slope}
\end{table}

\subsection{Run-Time Analysis- Analytical vs Automatic Differentiation}
 \label{sec:ID_SO_AD}

\begin{figure}
 \centering
 \includegraphics[width=1.0 \columnwidth]{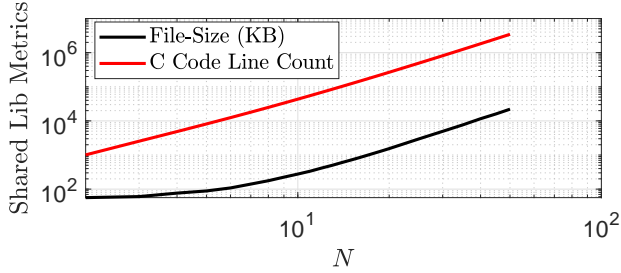}
 \caption{File size (in KB) and C code line count for ID SO AD derivatives computed using CasADi implemented in C++ Pinocchio for serial chains. The C code size increases with $N$, which leads to a large shared library size.}
 \label{ID_SO_AD_file_info}
 \end{figure}

To get the SO derivatives of Inverse Dynamics using AD, the IDSVA FO algorithm~\cite{singh2022efficient} is differentiated using the CasADi~\cite{andersson2012casadi} toolbox in C/C++ and the Pinocchio library. CasADi symbolic functions for the quantities $\frac{\partial \taubar}{\partial \q}$, $\frac{\partial \taubar}{ \partial \qd}$, and $\frac{\partial \taubar}{\partial \qdd}$ are differentiated by using CasADi's ``Jacobian'' functionality, to get the expressions for $\frac{\partial^2 \taubar}{\partial \q^2}$, $\frac{\partial^2 \taubar}{\partial \qd^2}$, $\frac{\partial^2 \taubar}{\partial \q\partial \qd}$, and $\frac{\partial^2 \taubar}{\partial \q \partial \qdd}$. Then the C code for each of the terms for any specific robot model is generated using the code-generation (CodeGen) capability in CasADi. For better run-time performance, the C code is compiled into a shared library, using the GCC 9.4.0 compiler with \texttt{O3} and \texttt{march=native} optimization flags. Although the process is straightforward to set up, for models with $N>30$ the AD approach results in  significant C code and large memory storage for the generated shared library. The code and library generation process is also time-consuming, taking as much as a few hours for a 36 DoF ATLAS humanoid on a laboratory desktop. Figure~\ref{ID_SO_AD_file_info} shows the C code line count and the memory storage of the shared libraries for the SO derivatives of Inverse Dynamics using AD, for serial chains up to $N=50$. For $N=40$, the C code is on the order of millions of lines of code, and the library size is $\approx$12 Mbs in size.

\begin{figure}
 \includegraphics[width=1.0 \columnwidth]{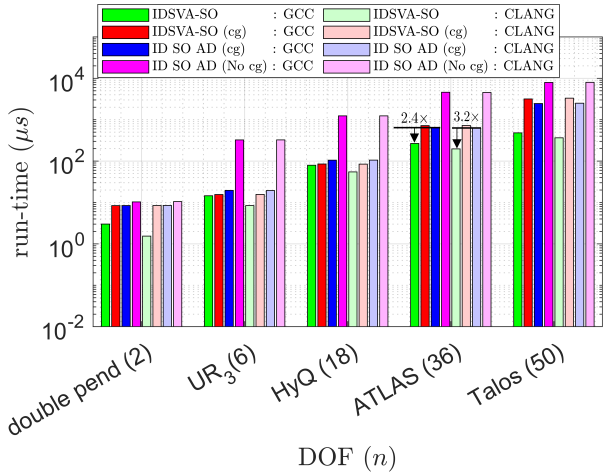}
 \caption{IDSVA-SO algorithm outperforms IDSVA-SO + CodeGen and ID SO using CasADi integrated with Pinocchio~\cite{carpentier2019pinocchio} C++ library. Two compilers are used- GCC (dark colors), and Clang (light colors). Fixed base models- double pendulum, $\rm{UR_3}$, floating base: Quadruped- HyQ, Humanoids- ATLAS, Talos.}
 \label{ID_SO_pin_AD_vs_ana}
 \end{figure}

The run-time comparison for several fixed-base and floating-base models including quadrupeds and humanoids for the AD approach with the IDSVA-SO algorithm is given in Fig.~\ref{ID_SO_pin_AD_vs_ana}. The IDSVA-SO algorithm is implemented in the C++ library Pinocchio~\cite{carpentier2019pinocchio}, and an open-source version is provided at \cite{cppsource}. AD approach without the CodeGen (in magenta) is also shown to highlight the usefulness of CodeGen. All of the algorithms are compiled using GCC 9.4.0 and CLANG 10.0 compilers in Ubuntu. Due to the use of the optimized shared library, the CodeGen AD approach is between $3-17 \times$ faster than the former. A CodeGen version of the IDSVA-SO algorithm (in red) in the CasADi framework is also provided. As seen from Fig.~\ref{ID_SO_pin_AD_vs_ana}, for $N<20$, the IDSVA-SO with (red) and without (green) CodeGen perform similarly, but the performance of IDSVA-SO+ CodeGen worsens at higher $N$ due to large C code and associated overhead. Since the IDSVA-SO algorithm is implemented as model-based, CodeGen ideally results in even more compiler optimizations for a particular model. However, the lack of benefits to using the CodeGen approach on IDSVA-SO at higher $N$ ($N>36$),  shows the fine-tuning of the IDSVA-SO algorithm implementation. As compared to the AD CodeGen approach, the IDSVA-SO is between $1.2-5 \times$ faster, and the speed-ups increase with $N$. For a 36 DoF ATLAS humanoid, the IDSVA-SO is approximately $3.2 \times$ faster than the AD approach upon using the CLANG compiler.
 
\subsection{Accuracy of Derivatives}
\label{sec:ID_SO_error}
Several error metrics are used to compare the accuracy of the derivatives calculated by the analytical and the alternative methods described in Sec.~\ref{sec:ID_SO_theory} and \ref{sec:ID_SO_alternative}. The Maximum Absolute Error (MAE) simply calculates the maximum value of the difference tensor and is defined as:

\begin{equation}
    e_{\rm max,abs} = \max|\mathcal{A}-\mathcal{A}_{\rm ref}| = ||\mathcal{A}-\mathcal{A}_{\rm ref}||_\infty
    \label{max_abs_error}
\end{equation}

Here $\mathcal{A}$ is the ID SO tensor computed using the method for which the accuracy is evaluated, while $\mathcal{A}_{\rm ref}$ is the reference tensor, also called ``truth''. Since the Bi-complex method is accurate to machine precision, it is used as the reference for computing the error metrics for the rest of the methods. The Root-Mean-Square Absolute Error (RMSAE) takes the root mean square of the component by component difference:
\begin{equation}
    e_{\rm rms,abs} = \sqrt{\frac{\sum_{i}\sum_{j}\sum_{k}|\mathcal{A}_{ijk}-A_{{\rm ref},ijk}|^2}{9n^3}}
    \label{rms_abs_error}
\end{equation}
Since the dimension of the ID SO tensor ($\mathcal{A}$) is $n \times 3n \times 3n$, the total number of elements used for RMSAE is $9n^3$. Along with the absolute difference, relative error is also considered. The Maximum Relative Error (MRE) is defined as:
\begin{equation}
    e_{\rm max,rel} = \max_{ijk} \Bigg|\frac{\mathcal{A}_{ijk}-\mathcal{A}_{{\rm ref},ijk}}{ \max(|\mathcal{A}_{{\rm ref},ijk}|,1)}\Bigg|
    \label{max_rel_error}
\end{equation}
Here, the component-by-component relative difference is computed for the ID SO tensor and then the maximum of those represents the error. The root mean square version of relative error (RMSRE) is defined as:
\begin{equation}
    e_{\rm rms,rel} = \sqrt{\bigg(\frac{1}{9n^3}\bigg)\sum_{i}\sum_{j}\sum_{k}\bigg(\frac{\mathcal{A}_{ijk}-\mathcal{A}_{{\rm ref},ijk}}{\max(|\mathcal{A}_{{\rm ref},ijk}|,1)} \bigg)^2}
    \label{rms_rel_error}
\end{equation}

\begin{figure}
 \includegraphics[width=1.0 \columnwidth]{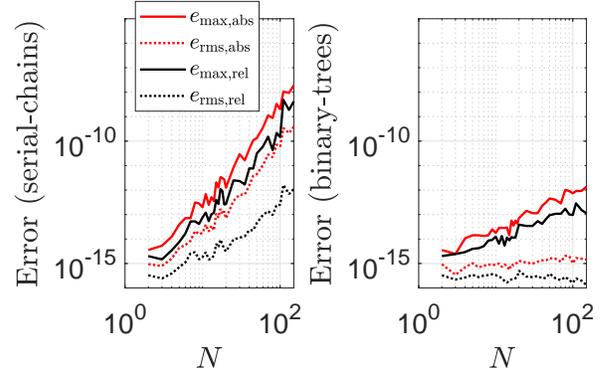}
 \caption{Error metrics for absolute and relative errors for the ID SO derivatives for serial chains and binary trees using the IDSVA-SO analytical method. The Bi-Complex derivatives are used as the reference for truth.}
 \label{ID_SO_error_v1}
 \end{figure}

Both of the relative error metrics give information about the number of significant digits preserved, especially if the derivative values are larger than 1. Figure~\ref{ID_SO_error_v1} shows the four error metrics explained for the ID SO derivatives for serial chains and binary trees computed using the IDSVA-SO algorithm, implemented in Fortran 90. As expected, the error metrics based on maximum value report higher errors as compared to the RMS errors. Since the RMS errors amplify the larger effects in errors, their smaller values show the high accuracy of the IDSVA-SO algorithm. The increase in errors with $N$ is expected due to the recursive nature of the algorithms. For the serial-chains, the errors accumulate as the size of the chain increases. This effect can be seen by the RMS error metrics. The binary trees have smaller chains along which the errors can accumulate, leading to smaller RMS error metrics. 

  \begin{figure}
 \includegraphics[width=1.0 \columnwidth]{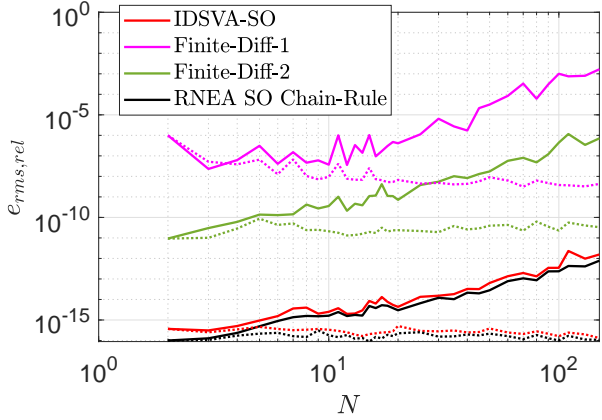}
 \caption{Comparison of error metrics for various methods to get ID SO derivatives for serial chains (bold lines) and binary trees (dotted lines) in Fortran 90. }
 \label{ID_SO_error_v2}
 \end{figure}
 
Figure~\ref{ID_SO_error_v2} provides the RMSRE \eqref{rms_rel_error} metric for the various alternative methods explained in Sec.~\ref{sec:ID_SO_alternative} and IDSVA-SO method, for serial chains and binary trees. Since the Chain-Rule method and IDSVA-SO are both analytical, they result in similar error metrics. As expected, the Finite-Diff-2 performs much better than Finite-Diff-1 by exploiting the analytical IDSVA as an intermediate step. For a binary tree with $N=40$, the difference in RMSRE between the two finite difference methods is $\approx10^{-3}$. From Fig.~\ref{ID_SO_error_v2}, the errors for the binary trees are on average lower than that of serial chains, since for the binary trees, many of the derivatives are zero due to the branch-induced sparsity. This results in sparse tensors for the binary trees, as compared to dense tensors for serial chains.

The previous two sections developed analytical expressions for the ID SO derivatives, followed by run-time and accuracy comparisons of the algorithm with several other techniques. The next section describes how these ID SO derivatives are exploited to compute the FD SO derivatives efficiently.

\section{SO Derivatives of Forward Dynamics: Theory} 
\label{sec:FD_SO_theory}

 
 \begin{figure}
 \centering
 \includegraphics[width= \columnwidth]{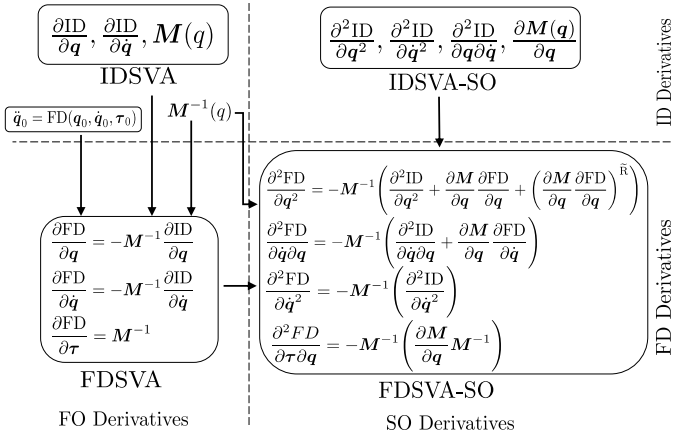}
  \caption{Computational flow for the Forward Dynamics derivatives. The top and down halves are for Inverse and Forward Dynamics derivatives, respectively, while the left and right halves are for the FO and the SO derivatives, respectively }
 \label{derivatives_flow}
 \end{figure}
 
  Similar to the ID SO derivatives, some of the SO derivative expressions for the $\textrm{FD}(\q,\qd,\taubar)$ are also trivial. From \eqref{fwd_dyn2}, since $\frac{\partial \textrm{FD}}{\partial \taubar} = \M^{-1}(\q)$, $\frac{\partial^2 \textrm{FD}}{\partial \taubar \partial \qd} = \frac{\partial^2 \textrm{FD}}{\partial \taubar^2}={\bf 0}$, while $\frac{\partial^2 \textrm{FD}}{\partial \taubar \partial \q}=\frac{\partial \M^{-1}}{\partial \q}$. The identity \cite{carpentier2018analytical} 
      \begin{equation}
        \frac{\partial \M^{-1}}{\partial \q} = -\M^{-1}\frac{\partial \M}{\partial \q} 
        \M^{-1}
        \label{sandwich_iden}
    \end{equation} 
     allows us to reuse the quantities $\M^{-1}$ and  $\frac{\partial\M}{\partial \q}$, which are already computed to get the FO derivatives and the ID SO derivatives, respectively. This approach is also shown in Fig.~\ref{derivatives_flow}, which depicts the computational flow of the FO and SO derivatives of Inverse $\&$ Forward Dynamics. The FD SO derivatives (at the bottom right of the flow chart) re-use the common quantities like $\M^{-1}$, and FD FO derivatives, which can be computed as follows.

The FO partial derivatives of FD can be written by exploiting the derivatives of ID as~\cite{jain1993linearization,carpentier2018analytical}:
  \begin{equation}
      \frac{\partial~\textrm{FD}}{\partial \boldsymbol u}\biggr\rvert_{\q_{0},  \qd_{0}, \taubar_{0}} = -\M^{-1}(\q_{0}) \frac{\partial~  \textrm{ID} }{\partial \boldsymbol u}\biggr\rvert_{\q_{0},  \qd_{0},  \qdd_{0}}
      \label{jain_FO_eqn}
  \end{equation}
  where $\u, \w \in \{\q,\qd\}$. To get the FD SO derivatives w.r.t $\q, \qd$,  an extension of ~\eqref{jain_FO_eqn} to get the SO derivatives of FD was provided by Ref.~\cite[Eq.~15]{nganga2021accelerating} for scalar variables. This is extended to handle the vector quantities $\u, \w$ as:

\begin{equation}
    \frac{\partial^2 \textrm{FD} }{\partial \u \partial \w} = \underbrace{-\M^{-1}\Bigg(}_\textit{Outer-Term} \underbrace{\frac{\partial^2 \textrm{ID}}{\partial \u \partial \w} + \frac{\partial \M}{\partial \w}\frac{\partial \textrm{FD}}{\partial \u} + \bigg(\frac{\partial \M}{\partial \u}\frac{\partial \textrm{FD}}{\partial \w}\bigg)\Rten}_\textit{Inner-Term} \Bigg)
    \label{ipr_so_1}
\end{equation}
where, $\w,\u \in \{\q,\qd\}$. For a particular $\w$ and $\u$:

\begin{enumerate}
    \item $\w,\u = \{\q\}$:
    In this case, the expression stays in its original form.
\item $\w,\u = \{\qd\}$: For this case, since $\M(\q)$ doesn't depend on $\qd$, the second and the third inner-terms in ~\eqref{ipr_so_1} vanish:

\begin{equation}
    \frac{\partial^2 \textrm{FD} }{\partial \u \partial \w} = -\M^{-1}\Bigg( \frac{\partial^2 \textrm{ID}}{\partial \u \partial \w} \Bigg) 
        \label{ipr_so_case2}
\end{equation}

\item $\w = \{\q\} $, $\u = \{\qd \}$:
For this case, only the third term in ~\eqref{ipr_so_1} vanishes:

\begin{equation}
    \frac{\partial^2 \textrm{FD} }{\partial \u \partial \w} = -\M^{-1} \Bigg( \frac{\partial^2 \textrm{ID}}{\partial \u \partial \w} + \frac{\partial \M}{\partial \w}\frac{\partial \textrm{FD}}{\partial \u} \Bigg)
    \label{ipr_so_case3}
\end{equation}

\item $\w = \{\qd\} $, $\u = \{\q \}$: For this case, only the second term in ~\eqref{ipr_so_1} vanishes:

\begin{equation}
    \frac{\partial^2 \textrm{FD} }{\partial \u \partial \w} = -\M^{-1} \Bigg( \frac{\partial^2 \textrm{ID}}{\partial \u \partial \w} + \bigg( \frac{\partial \M}{\partial \u}\frac{\partial \textrm{FD}}{\partial \w}\bigg)\Rten \Bigg)
    \label{ipr_so_case4}
\end{equation}

\end{enumerate}



For a scalar variable $\w$ and $\u$, \eqref{ipr_so_1} results in an $n$-vector quantity. When considering multi-DoF joints and a floating base, the results can be 3D tensor blocks. 
For clarity, the quantity inside the parenthesis is collectively called the \textit{Inner-Term}, and the quantity resulting from the product of $\M^{-1}$ and the \textit{Inner-Term} is called the \textit{Outer-Term}. For $\u,\w = [\q,\qd]$, since the second and the third quantities for \textit{Inner-Term} result in a $n \times n \times 2n$ tensor $\big(\frac{\partial \M}{\partial \w} \big)$ and a $n \times 2n$ matrix $\big(\frac{\partial \rm{FD}}{\partial \u}\big)$ product, the computational complexity of carrying out this the matrix/tensor product scales with an order $\mathcal{O}(N^4)$. Similarly, for the \textit{Outer-Term} too, the product of $\M^{-1}$ with \textit{Inner-Term} also results in a complexity of $\mathcal{O}(N^4)$. This $\mathcal{O}(N^4)$ Direct-Tensor-Matrix (DTM) approach gets expensive, especially if $N$ is high. More efficient alternative techniques for computing the \textit{Inner-Term} and \textit{Outer-Term} are presented in the following section. Our method for computing the SO derivatives in \eqref{ipr_so_1} by using FO analytical derivatives of FD (i.e., using FDSVA~\cite{singh2022efficient}), SO analytical derivatives of ID~(Sec.~\ref{sec:ID_SO_theory}), and analytical $\M^{-1}$~\cite{carpentier2018analyticalMinv} is called as the \textbf{FDSVA-SO} algorithm, also depicted in the flow chart in Fig.~\ref{derivatives_flow}. 

\subsection{Efficient Implementation Techniques}
\label{sec:eff_impl}
For an efficient algorithm implementation of \eqref{jain_FO_eqn}, Ref.~\cite{singh2022analytical} suggested the ABA-Zero-Algorithm (AZA), where the product of $\M^{-1}$ with any $n$-vector $\b$ can be computed by exploiting the form of the forward dyanmics:
\begin{equation}
 \M^{-1}(\q)\left( \taubar -  \C(\q,\qd)\qd - \g(\q) \right)= ABA(\q,\qd,\taubar,\g) \\
    \label{AZA_eqn}
\end{equation}
%
By running ABA with zero $\qd$, zero generalized gravity $\g$, and by inputting $\b$ instead of $\taubar$, the result $\M^{-1} \b$ is obtained. With this approach, the usual $\mathcal{O}(N^3)$ Direct-Matrix-Multiply (DMM) for implementing the FO Forward-Dynamics derivatives \eqref{jain_FO_eqn} can replaced by the efficient $n$ calls to the $\mathcal{O}(N)$ AZA approach. This method results in computational savings when $N>50$~\cite{singh2022efficient}. 

Instead of the expensive $\mathcal{O}(N^4)$ DTM approach to get the \textit{Outer-Term} in \eqref{ipr_so_1}, AZA can be also be used for each column of the 3D tensor \textit{Inner-Term}, thus resulting in a method with complexity $\mathcal{O}(N^3)$. 

To reduce the complexity of the \textit{Inner-Term}, the IDSVA algorithm~\cite[Alg. 1]{singh2022analytical} is exploited. IDSVA returns $\frac{\partial \taubar}{\partial \u}$, given $\q$,$\qd$,$\qdd$, and using \eqref{inv_dyn}, can be expressed as:
\begin{equation}
    \frac{\partial \taubar}{\partial \u} = \frac{\partial \M}{\partial \u}\qdd + \frac{\partial [\C \qd]}{\partial \u } + \frac{\partial \g}{\partial \u} = \textrm{IDSVA}(\q,\qd,\qdd,\g)
    \label{IDSVA_eqn}
\end{equation}
For zero $\qd$ and $\g$, and inputting an arbitrary $n$-vector $\b$ instead of $\qdd$, IDSVA returns the product of $\frac{\partial \M}{\partial \u}$ with $\b$. This variant of IDSVA shown in \eqref{IDFOZA_eqn}, is called \underline{\bf{I}}nv. \underline{\bf{D}}yn. \underline{\bf{F}}irst \underline{\bf{O}}rder \underline{\bf{Z}}ero \underline{\bf{A}}lgorithm, or IDFOZA.
\begin{equation}
    \frac{\partial \M}{\partial \u}\b = \textrm{IDSVA}(\q,\bf{0},\b, \bf{0})
        \label{IDFOZA_eqn}
\end{equation}
To get the second and third quantities in the \textit{Inner-Term}, IDFOZA is used for each column of the matrix $\frac{\partial FD}{\partial \u}$ and $\frac{\partial FD}{\partial \w}$, one at a time. Since IDSVA is an $\mathcal{O}(Nd)$ algorithm, the IDFOZA approach results in a complexity of $\mathcal{O}(N^2d)$, as compared to $\mathcal{O}(N^4)$ complexity of DTM.

\begin{figure}
 \centering
 \includegraphics[width=1.0 \columnwidth]{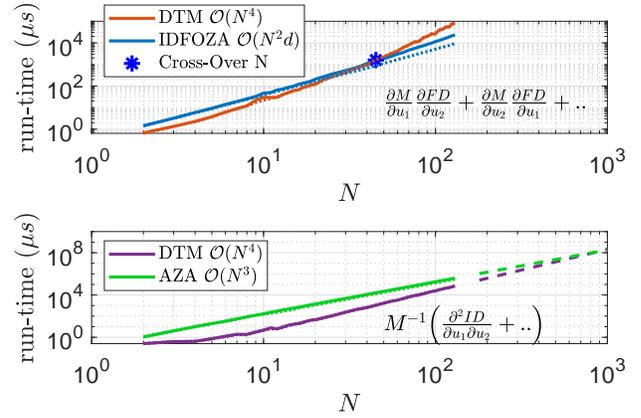}
 \caption{Run-time comparison for a) \textit{Inner-Term} computation using DTM and IDFOZA. The Cross-Over for \textit{Inner-Term} lies at $N=45$ for serial chains, and $N=25$ for binary trees.  b) \textit{Outer-Term} using DTM and AZA, for serial chains (bold line), and binary trees (dotted line). Dashed curves show linear interpolation of run-time data using {\sc Matlab} {\sc polyfit} function.} 
 \label{DTM_AZA_fortran}
 \end{figure}

 Fig.~\ref{DTM_AZA_fortran} shows the run-time comparison (in $\mu s$) for computing the \textit{Inner-Term} and \textit{Outer-Term} for serial chains and binary trees (branching factor $bf=2$) with the techniques explained above using Fortran 90. For the \textit{Inner-Term}, there is a  ``Cross-Over'' in performance at $N=45$ for serial-chains, beyond which the $\mathcal{O}(N^2d)$ IDFOZA approach is more efficient than the $\mathcal{O}(N^4)$ DTM approach. It is clear that for the \textit{Outer-Term}, the ``Cross-Over'' $N$ lies beyond $N=100$, and is of no use for practical robotics models of interest, resulting in DTM being the preferred approach in \eqref{ipr_so_1}. 
 

\begin{figure}
 \centering
 \includegraphics[width=1.0 \columnwidth]{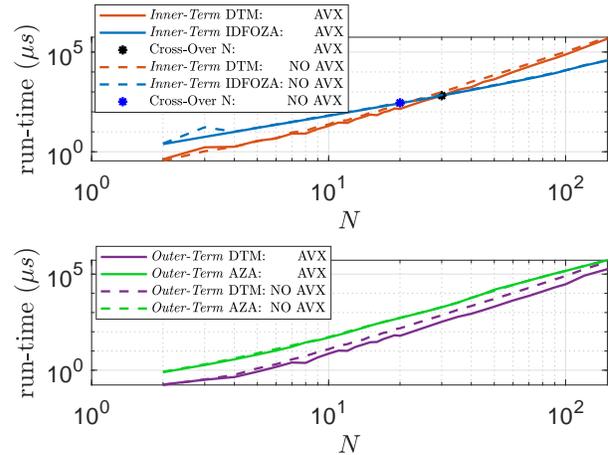}
 \caption{Run-Time comparison for \textit{Inner-Term} and \textit{Outer-Term} using DTM and smart implementation techniques for serial chains in C++ using the GCC 9.4.0 compiler. a) For \textit{Inner-Term}, the Cross-Over $N$ ``with'' AVX setting lies at $N=35$, and ``without'' AVX setting lies at $N=25$}. 
 \label{DTM_AZA_C}
 \end{figure}

The algorithms introduced in Sec.~\ref{sec:eff_impl} were also implemented in the C/C++ library Pinocchio~\cite{carpentier2019pinocchio}, and compiled using the GCC 9.4.0 compiler. Figure~\ref{DTM_AZA_C} shows a comparison of the methods used for computing the \textit{Inner-Term} and \textit{Outer-Term} for serial chains. To get better run-time performance, the compiler optimization flag \texttt{O3}, along with \texttt{march=native} is used. This triggers the processor (12th Gen Intel i5-12400) to use the Advanced Vector Extensions (AVX-2), and thus boost the run-time performance for matrix and vector products, resulting in lower run-time for the DTM approach. This benefit of using AVX for the DTM technique is shown in Fig.~\ref{DTM_AZA_C}. For the \textit{Inner-Term}, since the run-time curve for DTM (denoted by solid brown) shifts down, the Cross-Over $N$ with AVX (black asterisk) lies further down the curve, as compared to without using AVX (blue asterisk). For the \textit{Outer-Term}, the use of AVX for DTM results in a speedup between $1.05-3 \times$ for serial chains. Cross-Over values for the C++ implementation are slightly different from the ones obtained by the Fortran version, given in Fig.~\ref{DTM_AZA_fortran}. Hence, the choice of algorithm to get the best run-time performance for the FD SO derivatives depends on the hardware, implementation environment, and compiler optimization settings.
  

\section{SO Derivatives of Forward Dynamics: Results}
\label{sec:FD_SO_results}

 \subsection{Run-Time Analysis - Analytical vs Complex-Step, Chain-Rule, \& Finite-Difference}
 \label{sec:FD_SO_results1}

\begin{figure}
 \centering
 \includegraphics[width=1.0 \columnwidth]{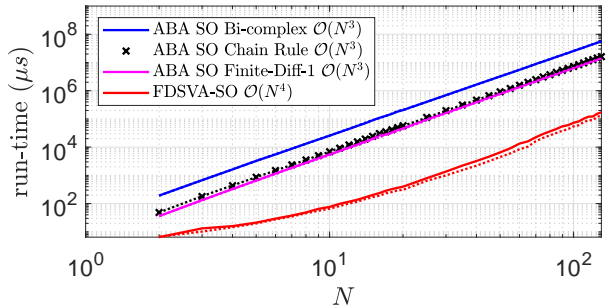}
 \caption{Forward Dynamics SO Derivatives run-times for serial chains (bold line, crossed line), and binary trees (dotted line). FDSVA-SO outperforms the alternative methods for serial chains and binary trees for all $N$.} 
 \label{FD_SO_full_fortran}
 \end{figure}

\begin{figure}
 \centering
 \includegraphics[width=1.0 \columnwidth]{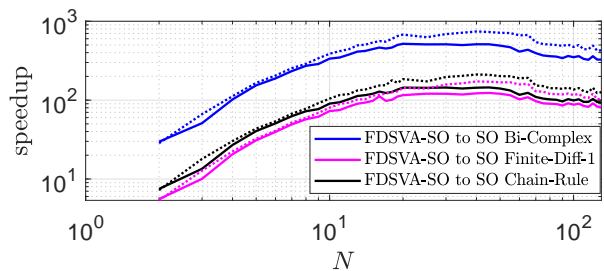}
 \caption{The speedups of FDSVA-SO over the alternative methods are higher for binary trees (dotted lines) than serial chains (bold lines) due to the branch-induced sparsity exploited by the analytical FDSVA-SO method. } 
 \label{FD_SO_full_fortran_speedup}
 \end{figure}

\begin{figure}
 \centering
 \includegraphics[width=1.0 \columnwidth]{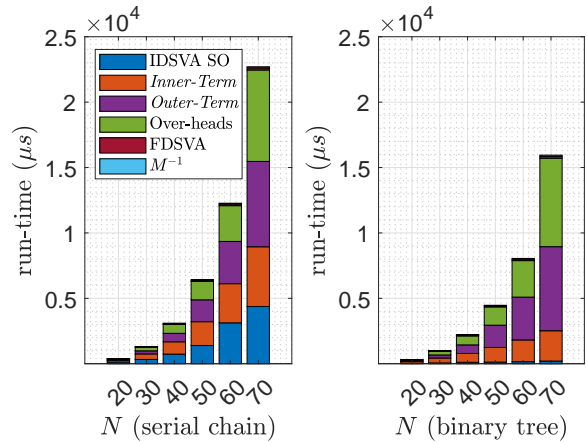}
 \caption{Breakdown of the FDSVA-SO timings for serial chains and binary trees. Due to branch-induced sparsity, the run times for binary trees for IDSVA-SO and \textit{Inner-Term} are much lesser as compared to serial trees.} 
 \label{FD_SO_Fortran_breakdown}
 \end{figure}

 Figure~\ref{FD_SO_full_fortran} compares the run-times for various alternative methods introduced in Sec.~\ref{sec:ID_SO_results} with the analytical FDSVA-SO method (Sec.~\ref{sec:FD_SO_theory}) for serial and binary trees, implemented in Fortran 90. Since each of these alternative methods computes SO derivatives of the $\mathcal{O}(N)$ ABA algorithm, they result in computational complexity of $\mathcal{O}(N^3)$. The Finite-Difference method used here is the Finite-Diff-1 approach (see App.~\ref{sec:finite_diff1}) around the ABA algorithm. The run-times for Finite-Diff-2 (see App.~\ref{sec:finite_diff2}) approach which uses a FO central difference formula around FDSVA~\cite{singh2022analytical} is skipped here but could prove better in terms of run-time and accuracy as compared to Finite-Diff-1 approach. The SO chain rule proves more expensive than Finite-Difference, unlike the case in ID-SO derivatives (Fig.~\ref{fort_ID_SO_v1}). This flip in performance is due to ABA having more intermediate computations as compared to RNEA, due to an extra forward pass. By comparison to Chain rule methods, the $\mathcal{O}(N^4)$ analytical FDSVA-SO method shows significant speedups between $5-500\times$, as shown in Fig.~\ref{FD_SO_full_fortran_speedup}.
 The breakdown of the run-time for FDSVA-SO for serial and binary tees is given in Fig.~\ref{FD_SO_Fortran_breakdown}. Below $N<40$, the \textit{Inner-Term} timings are higher than the \textit{Outer-Term} timings. But with an increase in $N$, the \textit{Outer-Term} becomes more and more expensive due to the $\mathcal{O}(N^4)$ computations, as shown in Fig.~\ref{DTM_AZA_fortran}. The over-heads include the slicing and assignment operations in the resulting 3D tensor. Due to the structure of the tensor, the memory accesses are non-contiguous, resulting in a high number of cache misses, and hence high overheads at high $N$. Lastly, the run times for IDSVA-SO for the binary trees are improved, as expected, compared to the serial trees due to the branch-induced sparsity.
 
 \subsection{Run-Time Analysis: Analytical vs Automatic Differentiation}
 \label{sec:FD_SO_c}
\begin{figure}
 \centering
 \includegraphics[width=1.0 \columnwidth]{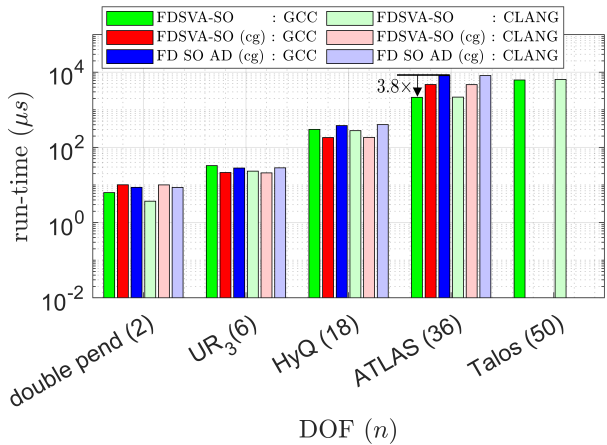}
 \caption{FD SO Derivatives - Comparison of Analytical FDSVA-SO, FDSVA-SO + CodeGen, AD (CasADi) + CodeGen. ``cg'' denotes CodeGen. For the ATLAS humanoid, FDSVA-SO is approximately $3.8 \times$ faster than the AD+CodeGen approach using the GCC compiler.} 
 \label{FD_SO_pin_AD_vs_ana}
 \end{figure}

 Similar to the ID SO derivatives, the FD SO derivatives are also computed using the AD approach. Following the method in Sec.~\ref{sec:ID_SO_AD}, the FDSVA~\cite{singh2022efficient} is differentiated using the CasADi toolbox to get the FD SO derivatives. Figure~\ref{FD_SO_pin_AD_vs_ana} shows a comparison of the FDSVA-SO algorithm with the FD SO derivatives using the AD approach. A CodeGen version (red) of the FDSVA-SO algorithm in CasADi is also compared. All of the algorithms are implemented in the Pinocchio C++ library and compiled using the GCC 9.4.0 and Clang 10.0 compilers. For the 6-DoF {$\rm{UR_{3}}$} and 18-DoF HyQ, FDSVA-SO with CodeGen outperforms the normal FDSVA-SO algorithm, highlighting the benefit of the CodeGen approach to optimize the algorithms for a particular model. The strong CodeGen performance is also due to the inefficient slicing tensor operations in the Eigen C++ library that are key to the FDSVA-SO algorithms. But at $n=36$ and higher, the CodeGen process becomes time and memory-consuming due to the large C code being generated, making the analytical method the preferred choice. The C code file size resulting from the CodeGen method for the Talos humanoid model with $n=50$ is $\approx 500$ megabytes, making it impossible to generate the shared files. Hence, the FDSVA-SO method is the only possible choice here.

\subsection{Accuracy of Derivatives}
\begin{figure}
\includegraphics[width=1.0 \columnwidth]{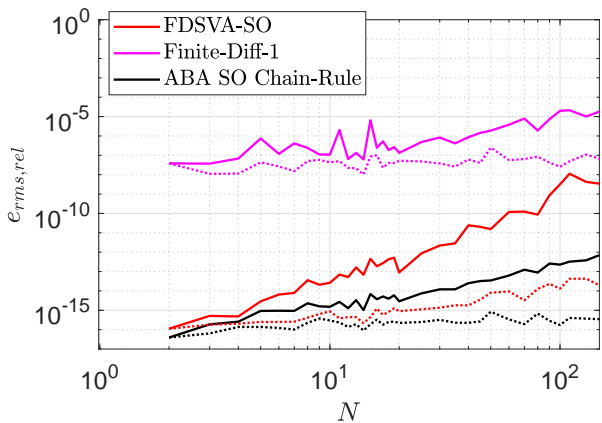}
\caption{Accuracy analysis (RMSRE) for FD SO derivatives for serial chains (bold lines) and binary trees (dotted lines) in Fortran 90. The ABA SO Chain-Rule performs significantly better than the FDSVA-SO as $N$ increases.}
\label{FD_SO_error}
\end{figure}

Similar to the accuracy analysis shown in Sec.~\ref{sec:ID_SO_error}, the four error metrics are also computed for the FD SO derivatives for the methods explained in Sec.~\ref{sec:FD_SO_results1}. The FD SO derivatives computed using the Bi-complex method over the ABA algorithm is used again as the reference for the error metrics. Figure~\ref{FD_SO_error} shows a comparison of RMSRE for the FDSVA-SO, the Finite-Diff-1, and the Chain-Rule method over ABA. Similar to the analysis shown in Fig.~\ref{ID_SO_error_v2}, the Finite-Diff-1 performs the worst in terms of accuracy. Since the FDSVA-SO comprises multiple steps from different algorithms (see Fig.~\ref{derivatives_flow}), the resulting errors accumulate. For a serial chain with $N=40$, the FDSVA-SO has RMSRE of $\approx 10^{-11}$, as compared to $\approx 10^{-14}$ for the Chain-Rule method. However, the RMSRE values are bounded by $10^{-8}$ for the FDSVA-SO method (for $N<150$), proving it to have an accuracy that should be acceptable for applications. As the case for ID SO derivatives too, the error for the binary trees is lower than the serial chains, in part due to the presence of shorter chains, with smaller derivative values. This instead leads to smaller precision loss and hence better accuracy.

\section{Conclusion}
In this work, simplified analytical expressions are presented for the second-order derivatives of Inverse and Forward rigid-body dynamics for models with multi-DoF joints and fixed or floating bases. A recursive algorithm for the second-order derivatives of Inverse Dynamics is presented, and run-time comparisons are shown with alternative methods like the Bi-Complex step, Finite-Difference, Chain-Rule accumulation, and Automatic Differentiation (AD). Efficient implementation tricks are also presented to compute the second-order derivatives of Forward Dynamics by exploiting the first-order derivatives presented in previous works. The benefits of using code-generation techniques using CasADi to speed up the implementation of the AD and the analytical method are shown for some of the cases. Finally, MATLAB and C++ versions of the presented algorithms are open-sourced for use in trajectory optimization algorithms like DDP and SQP. Future extensions of this work can focus on developing parallel computing algorithms (e.g., for use with GPUs), building on the FO case in Ref.~\cite{plancher2021accelerating}. The derivatives developed herein can also be extended to get the derivatives for models with contact dynamics. 


\section*{Acknowledgments}
The authors thank Dr. Shivesh Kumar and Dr. Andreas M\"{u}ller for valuable discussions about the paper.

\appendices

\renewcommand{\thesubsection}{\Alph{section}.\Roman{subsection}}

\renewcommand{\thesubsectiondis}{\Roman{subsection}.}

\renewcommand{\thesubsubsection}{\Alph{section}.\Roman{subsection}.\arabic{subsubsection}}

\section{Finite-Difference \& Complex-Step}
\label{sec:alternative_method}
Various alternative methods used to compute the SO derivatives of ID and FD are detailed below:
 \subsection{Complex-Step}
 \label{sec:bi_complex}
The complex-step~\cite{squire1998using} method can be used to take accurate machine-precision derivatives of a function by perturbing the inputs in the complex plane. Considering a scalar multi-variable function $z=f(x,y)$:

\begin{equation}
    \frac{\partial z}{\partial x} = \frac{Im(f(x+ih,y))}{h}+\mathcal{O}(h^2)
\end{equation}
Here, $Im$ denotes the imaginary part of the perturbed function value, and $h$ is the step size. Since there is no precision loss in complex-step method due to subtractions, $h$ is chosen to a very small value of $10^{-20}$. A multi-complex extension~\cite{lantoine2012using} of the complex-step method called Multi-Complex-Step (MCX) allows to take the SO derivative as:
\begin{align}
    \frac{\partial^2 z}{\partial x^2} & = \frac{Im_{12}(f(x+i_{1}h+i_{2}h,y))}{h^2}+\mathcal{O}(h^2)\\ 
    \frac{\partial^2 z}{\partial x \partial y} &= \frac{Im_{12}(f(x+i_{1}h,y+i_{2}h))}{h^2} +\mathcal{O}(h^2)\nonumber
\end{align}
Here, $Im_{12}$ denotes the coefficient of the imaginary quantity $i_{1}i_{2}$ resulting from perturbing $f$ in the bi-complex plane, represented by a bi-complex number $a+i_{1}b+i_{2}c$. A vector-valued function can also be used instead of a scalar function.

\subsection{Finite-Difference Method}

The finite difference or ``Numerical-Differentiation'' is used to get the derivatives of a function or an algorithm by perturbing the inputs, one at a time. Although finite-differencing can be implemented in parallel, it suffers from poor accuracy. Two approaches to getting the SO derivatives of ID/FD using finite-difference are presented. 
 \subsubsection{Finite-Diff-1}
  \label{sec:finite_diff1}
The straightforward approach to getting the SO derivatives of ID is the second-order finite-difference of RNEA, and for the SO derivatives of Forward Dynamics, is the second-order finite-difference of ABA. Since the ABA and RNEA algorithms have a linear computational complexity ($\mathcal{O}(N)$), this finite-difference method results in a complexity of $\mathcal{O}(N^3)$. Considering quadratic truncation error for all derivatives, the central-difference method for a multi-variable function $z=f(x,y)$ is considered:

\begin{equation}
    \frac{\partial^2 z}{\partial x^2} = \frac{f(x^{+})-2f(x,y)+f(x^{-})}{h^2}+\mathcal{O}(h^2)
         \label{finite_diff1_eqn1}
\end{equation}

\begin{align}
 &\frac{\partial^2 z}{\partial x \partial y} = \frac{1}{4hk}\bigg(f(x^{+},y^{+})- f(x^{+},y^{-})-\nonumber \\ &~~~~~~~~~f(x^{-},y^{+})+ f(x^{-},y^{-})\bigg)+\mathcal{O}(h^2)+\mathcal{O}(k^2)
     \label{finite_diff1_eqn2}
\end{align}

Here, $f(x^{+})$ denotes a perturbation of function $f$ in positive $x$ direction, i.e. $f(x+h,y)$, and $f(x^{-}) = f(x-h,y)$. The quantity $f(x^{+},y^{+})$ denotes a perturbation in both the positive $x$ and $y$ direction as $f(x+h,y+k)$. The parameters $h$ and $k$ are the step sizes used for the finite-difference. In order to reduce the numerical error resulting from truncation and the round-off errors, an error analysis is done to find the optimal step size. The Root Mean Square Relative Error (RMSRE)(see \eqref{rms_rel_error}) is considered. 

\begin{figure}
 \centering
 \includegraphics[width=1.0 \columnwidth]{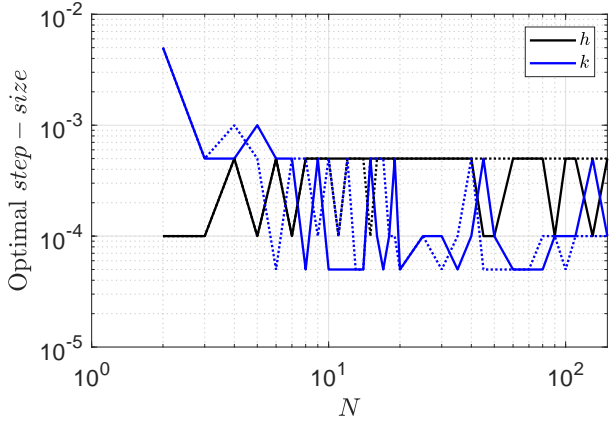}
 \caption{Step-sizes $h$ and $k$ for RNEA SO Finite-Diff-1 method for serial chains (bold), and binary trees (dotted), implemented in Fortran 90.}
 \label{ID_SO_finite_diff1_tuning}
 \end{figure}

\begin{figure}
 \centering
 \includegraphics[width=1.0 \columnwidth]{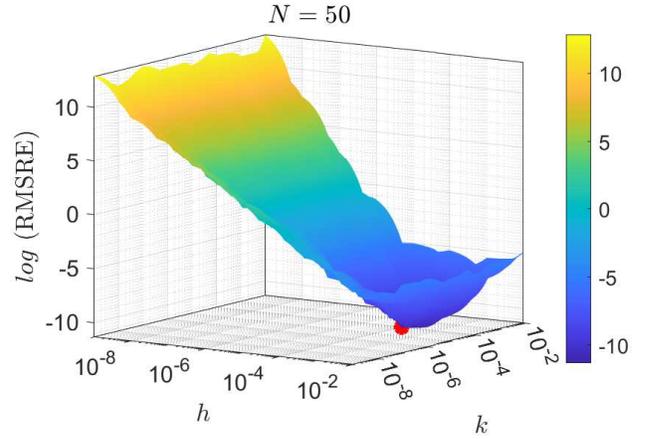}
 \caption{$\log$ (RMSRE) contours by varying the step-sizes $h$ and $k$ for Finite-Diff-1 method for ID SO derivatives for a serial chain with $N=50$, implemented in Fortran 90. The red dot denotes the min($\log$ (RMSRE)) value.}
 \label{ID_SO_finite_diff1_tuning_contour}
 \end{figure}

For ID SO derivatives, by varying the values of $h$ and $k$ within a pre-specified range, the RMSRE is calculated for serial chains and binary trees for $N$ between 2 and 110. The ``optimal'' values of the step-size $h$ and $k$ are the values of $h$ and $k$ for which the RMSRE is at the minimum.  Figure~\ref{ID_SO_finite_diff1_tuning} shows that these ``optimal'' values lie within  $10^{-4}$ and $10^{-3}$. Figure~\ref{ID_SO_finite_diff1_tuning_contour} shows the $\log (\textrm{RMSRE})$ contours by varying the step-sizes $h$ and $k$ for a serial chain with $N=50$.

 \subsubsection{Finite-Diff-2}
   \label{sec:finite_diff2}
\begin{figure}
 \centering
 \includegraphics[width=0.9 \columnwidth]{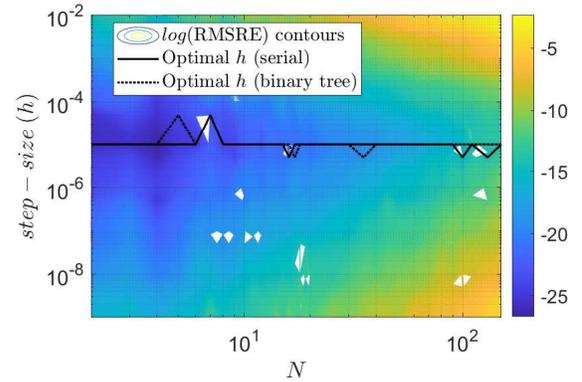}
 \caption{$\log$ (RMSRE) contours for Finite-Diff-2 approach to get the ID SO derivatives for serial chains (bold) ad binary trees (dotted), along with the ``optimal'' step-size $h$ value, implemented in Fortran 90. }
 \label{ID_SO_finite_diff2_tuning_contour}
 \end{figure}

Another method to compute the SO derivatives of a multi-variate function is to use the first-order central-difference on the FO derivative:
\begin{equation}
    \frac{\partial^2 z}{\partial x^2} = \frac{f^{'}(x+h,y)-f^{'}(x-h,y)}{2h}+\mathcal{O}(h^2)
    \label{finite_diff2_eqn}
\end{equation}
Here, the cross-derivative terms are computed directly by perturbing the opposite variable of the first derivative argument. For the ID SO derivatives, the FO IDSVA algorithm~\cite{singh2022efficient} is used as $f^{'}$ in \eqref{finite_diff2_eqn}, thus resulting in a hybrid analytical and finite-difference method. Compared to the two tuning parameters $h$ and $k$ for the Finite-Diff-1 approach, only one parameter ($h$) needs to be tuned for this method. Figure~\ref{ID_SO_finite_diff2_tuning_contour} shows the  $\log$(RMSRE) contours by varying the step-size $h$ to get the ID SO derivatives for serial chains and binary trees. With a few exceptions, for almost all $N$, the ``optimal'' step-size for this approach lies at $10^{-5}$. 
 
\section{Multi-DoF Joint Identities}
\label{mdof_iden}
The following spatial vector/matrix identities are derived by taking the partial derivative w.r.t the full joint configuration $(\q_{j})$, or joint velocity $(\qd_{j})$ of the joint $j$. In each, the
quantity to the left of the equals sign equals the
expression to the right if $j\preceq i$, and is zero otherwise,
unless otherwise stated.
The identities \ref{psii_dot} and \ref{Psiddoti} describe the partial derivatives of $\Psibardot_{i}$ and $\Psibarddot_{i}$ present in \eqref{dtau_dq}-\eqref{dtau_dqd}. Identities \ref{Iic}, \ref{Bic}, and \ref{fic} give the partial derivatives of the composite Inertia ($\IC{i}$), body-level Coriolis matrix ($\BC{i}$), and net composite spatial force on a body ($\f_{i}^{C}$). The quantities $\frac{\partial \I_{i}^{C}}{\partial \q_{j}}$(\ref{Iic}) and $\frac{\partial \B_{i}^{C}}{\partial \q_{j}}$ (\ref{Bic}) and  are third-order tensors where the partial derivatives of matrices $\BC{i}$ and $\IC{i}$ w.r.t each DoF of joint $j$ are stacked as matrices along the pages of the tensor. The tensor $\Bten{\I_{i}^{C}}{\Phibar_{j}}$ used in \ref{Bic_qdj} is also a composite with $\Phibar_{j}$ as the argument. The details of derivations for these identities can be found in Ref.~\cite{arxivss_SO}.
 
{
\small
\begin{align*}
&\frac{\partial \Phibar_{i}}{\partial \q_{j}}=  \Phibar_{j}   \timesM \Phibar_{i}\tag{K1}\label{dphii}\\
&\frac{\partial \Phibardot_{i}}{\partial \q_{j}}=
    \Psibardot_{j}   \timesM \Phibar_{i}+ \Phibar_{j} \timesM \Phibardot_{i}\tag{K2}\label{dphii_dot}\\
&\frac{\partial (\Phibar_{i}\qd_{i} \times \Phibar_{i})}{\partial \q_{j}}=
      \Phibar_{j}   \timesM (\Phibar_{i}\qd_{i} \times \Phibar_{i})\tag{K3}\label{vji_phii}\\
&\frac{\partial \Psibardot_{i}}{\partial \q_{j}} =
  \Psibardot_{j}   \timesM  \Phibar_{i} + \Phibar_{j} \timesM \Psibardot_{i}\tag{K4}\label{psii_dot}\\
&\frac{\partial \I_{i}}{\partial \q_{j}}=  
    \Phibar_{j} \timesfM \I_{i}- \I_{i}(\Phibar_{j} \timesM)\tag{K5}\label{Ii} \\
&\frac{\partial \I_{i}^{C}}{\partial \q_{j}}=
      \begin{cases}
      \Phibar_{j} \timesfM \I_{i}^{C}- \I_{i}^{C}(\Phibar_{j} \timesM), &~~~~ \text{if}\ j \preceq i \\
      \Phibar_{j} \timesfM \I_{j}^{C}- \I_{j}^{C}(\Phibar_{j} \timesM), &~~~~ \text{if}\ j \succ i 
    \end{cases} \tag{K6}\label{Iic}\\
& \frac{\partial \a_{i}}{\partial \q_{j}}= 
    \Psibarddot_{j} - \v_{i} \times \Psibardot_{j} - \a_{i} \times \Phibar_{j}\tag{K7}\label{ai} \\
&\frac{\partial (\I_{i}\a_{i})}{\partial \q_{j}}= 
    \Phibar_{j}\timesfM (\I_{i} \a_{i})+ \I_{i} \Psibarddot_{j} - \I_{i}( \v_{i} \times \Psibardot_{j})\tag{K8}\label{Iiai} \\
&\frac{\partial \Psibarddot_{i}}{\partial \q_{j}}=
  \Psibarddot_{j}   \timesM  \Phibar_{i} + 2\Psibardot_{j} \timesM \Psibardot_{i}+\Phibar_{j} \timesM \Psibarddot_{i}\tag{K9}  \label{Psiddoti}\\
&\frac{\partial \B_{i}^{C}}{\partial \q_{j}}=
     \begin{cases}
      \Bicpsiidot{i}{j}+\Phibar_{j} \timesfM \B_{i}^{C}- \B_{i}^{C}(\Phibar_{j} \timesM), & \text{if}\ j \preceq i \\
         \Bicpsiidot{j}{j}+\Phibar_{j} \timesfM \B_{j}^{C}- \B_{j}^{C}(\Phibar_{j} \timesM), & \text{if}\ j \succ i 
    \end{cases} \tag{K10}\label{Bic}\\ 
&\frac{\partial \f_{i}}{\partial \q_{j}}=
   \I_{i} \Psibarddot_{j} +  \f_{i}\crff  \Phibar_{j} + 2 \B_{i} \Psibardot_{j} \tag{K11}\label{fi} \\
&\frac{\partial \f_{i}^{C}}{\partial \q_{j}}=
      \begin{cases}
      \I_{i}^{C} \Psibarddot_{j} + \f_{i}^{C} \crff \Phibar_{j} + 2 \B_{i}^{C} \Psibardot_{j} ,   & \text{if}\ j \preceq i \\
      \I_{j}^{C} \Psibarddot_{j} + \f_{j}^{C}\crff\Phibar_{j} + 2 \B_{j}^{C} \Psibardot_{j}, &\text{if}\ j \succ i 
    \end{cases} \tag{K12}\label{fic}\\ 
&\frac{\partial \Phibar_{i} \T}{\partial \q_{j}}=
  -\Phibar_{i} \T\Phibar_{j} \timesfM\tag{K13}\label{PhiT} \\
&\frac{\partial \Phibardot_{i}}{\partial \qd_{j}}=
  \Phibar_{j} \timesM \Phibar_{i}\tag{K14} \label{Phidot_qdj} \\
&\frac{\partial \Psibardot_{i}}{\partial \qd_{j}}=
    \begin{cases}
         \Phibar_{j} \timesM \Phibar_{i}, &{\text{if}\ j \prec i} \\
      0,& \text{otherwise}
    \end{cases}\tag{K15}\label{Psidot_qdj} \\ 
&\frac{\partial \B_{i}^{C}}{\partial \qd_{j}}=
    \begin{cases}
      \Bten{\I_{i}^{C}}{\Phibar_{j}}, & \text{if}\ j \preceq i \\
    \Bten{\I_{j}^{C}}{\Phibar_{j}}, & \text{if}\ j \succ i 
    \end{cases} \tag{K16} \label{Bic_qdj} 
\end{align*}
}


\section{Summary}
\label{summary_sec}
\noindent\underline {\bf Common Terms:}
{\small %
\begin{align*}
  \red{\mathcal{A}_{1}} &\triangleq \red{\Phibar_{i} \timesfM \B_{i}^{C} - \B_{i}^{C}\Phibar_{i} \timesM}\\
    \blue{\mathcal{A}_{2}} & \triangleq \blue{\Phibar_{i} \timesfM \I_{i}^{C}   -\I_{i}^{C}   \Phibar_{i}\timesM}
\end{align*}
}

{\noindent \underline {\bf SO Partials w.r.t $\q$:}
{\small
\begin{align*}
\small	
&\frac{\partial^2 \taubar_{i}}{\partial \q_{j} \partial \q_{k}} =  -\left[2\Psibardot_{j}\T (\Bten{\I_{i}^{C}}{\Phibar_{i}}\Psibardot_{k})\Rten  + 2\Phibar_{j}\T ((\B_{i}^{C\T} \Phibar_{i}) \crffM\Psibardot_{k} )\Rten\right.  \\
&~~~~~~~~~~~~~~~~~~~~~~~~~~~~~\left.+\Phibar_{j}\T ((\I_{i}^{C}\Phibar_{i} ) \crffM\Psibarddot_{k} )\Rten \right]\Tten  , (k \preceq j \preceq i)\\
&\frac{\partial^2 \taubar_{i}}{\partial \q_{k} \partial \q_{j}} =  \left[\frac{\partial^2 \taubar_{i}}{\partial \q_{j} \partial \q_{k}} \right]\Rten, (k \prec j \preceq i)\\
&  \frac{\partial^2 \taubar_{k}}{\partial \q_{i} \partial \q_{j}}= \Phibar_{k}\T \Big[ \big[2(\Bten{\I_{i}^{C}}{\Psibardot_{i}}+\red{\mathcal{A}_{1}})\Psibardot_{j} + \blue{\mathcal{A}_{2}}\Psibarddot_{j} \big]\Rten +\\
& ~~~~~~~~~~~~~~~~\Phibar_{j} \timesfM \big( 2 \B_{i}^{C}  \Psibardot_{i} +\I_{i}^{C} \Psibarddot_{i}+\f_{i}^{C} \crff \Phibar_{i}\big) \Big]  , (k \prec j \preceq i)\\
&      \frac{\partial^2 \taubar_{k}}{\partial \q_{j} \partial \q_{i}} =  \left[\frac{\partial^2 \taubar_{k}}{\partial \q_{i} \partial \q_{j}} \right]\Rten, (k \prec j \prec i) \\
&  \frac{\partial^2 \taubar_{j}}{\partial \q_{k} \partial \q_{i}} = \Phibar_{j} \T \big[ 2(  \Bten{\I_{i}^{C}}{\Psibardot_{i}}+\red{\mathcal{A}_{1}}) \Psibardot_{k} +  \blue{\mathcal{A}_{2}} \Psibarddot_{k}\big] , (k \preceq j \prec i)\\
&    \frac{\partial^2 \taubar_{j}}{\partial \q_{i} \partial \q_{k}} =  \left[  \frac{\partial^2 \taubar_{j}}{\partial \q_{k} \partial \q_{i}} \right]\Rten, (k \preceq j \prec i)
\end{align*}
}

\noindent \underline {\bf SO Partials w.r.t $\qd$:}
{\small
\begin{align*}
&     \frac{\partial^2 \taubar_{i}}{\partial \qd_{j} \partial \qd_{k}} =-\big[\Phibar_{j}\T (2\Bten{\I_{i}^{C}}{\Phibar_{i}} \Phibar_{k})\Rten \big]\Tten, (k \prec j \preceq i)\\
&     \frac{\partial^2 \taubar_{i}}{\partial \qd_{k} \partial \qd_{j}} = \left[ \frac{\partial^2 \taubar_{i}}{\partial \qd_{j} \partial \qd_{k}} \right]\Rten, (k \prec j \preceq i)\\
&     \frac{\partial^2 \taubar_{i}}{\partial \qd_{j} \partial \qd_{k}} =   -\big[ \Phibar_{j}\T  (\blue{\mathcal{A}_{2}} \Phibar_{k} )\Rten\big]\Tten , (k = j \preceq i)\\
&  \frac{\partial^2 \taubar_{k}}{\partial \qd_{i} \partial \qd_{j}} =  \Phibar_{k}\T \Big[2\Bten{\I_{i}^{C}}{\Phibar_{i}} \Phibar_{j} \Big]\Rten, (k \prec j \prec i) \\
&     \frac{\partial^2 \taubar_{k}}{\partial \qd_{j} \partial \qd_{i}} = \left[  \frac{\partial^2 \taubar_{k}}{\partial \qd_{i} \partial \qd_{j}} \right]\Rten, (k \prec j \prec i)\\
&   \frac{\partial^2 \taubar_{k}}{\partial \qd_{i} \partial \qd_{j}} =  \Phibar_{k}\T \Big[ \big((\I_{i}^{C}\Phibar_{i})\crffM   +\Phibar_{i} \timesfM \I_{i}^{C}\big) \Phibar_{j}  \Big]\Rten , (k \prec j = i)\\
&      \frac{\partial^2 \taubar_{j}}{\partial \qd_{k} \partial \qd_{i}} = \Phibar_{j}\T \Big[2\Bten{\I_{i}^{C}}{\Phibar_{i}} \Phibar_{k}\Big], (k \preceq j \prec i)\\
&     \frac{\partial^2 \taubar_{j}}{\partial \qd_{i} \partial \qd_{k}} = \left[   \frac{\partial^2 \taubar_{j}}{\partial \qd_{k} \partial \qd_{i}}  \right]\Rten, (k \preceq j \prec i)
\end{align*}}
}

\noindent \underline {\bf Cross SO Partials w.r.t $\q$ and $\qd$:}
{\small
\begin{align*}
\small
&  \frac{\partial^2 \taubar_{i}}{ \partial \qd_{j}\partial \q_{k}} =    -\big[ \Phibar_{j}\T \big( 2\Bten{\I_{i}^{C}}{\Phibar_{i}} \Psibardot_{k}\big) \Rten  \big] \Tten , (k \preceq j \preceq i)\\
&    \frac{\partial^2 \taubar_{j}}{\partial \qd_{i} \partial \q_{k}} =    \Phibar_{j}\T \left[ 2\Bten{\I_{i}^{C}}{\Phibar_{i}}\Psibardot_{k}  \right] \Rten  , (k \preceq j \prec i)\\
&  \frac{\partial^2 \taubar_{i}}{\partial \qd_{k} \partial \q_{j}} =  \big[\Phibar_{k}\T (-2\Bten{\I_{i}^{C}}{\Phibar_{i}} \Psibardot_{j} + (2\B_{i}^{C\T} \Phibar_{i})\crffM \Phibar_{j} \\ 
&~+2(\I_{i}^{C}\Phibar_{i})\crffM \Psibardot_{j}  )\Rten 
+(\Psibardot_{k} + \Phibardot_{k} ) \T ((\I_{i}^{C}\Phibar_{i})\crffM \Phibar_{j}  )\Rten   
      \big]\Tten  \\
&~~~~~~~~~      ,(k \prec j \preceq i)\\
&  \frac{\partial^2 \taubar_{k}}{\partial \qd_{i} \partial \q_{j}} =   \Phibar_{k}\T \left[ \big(2 \B_{i}^{C}\Phibar_{i} +
  \I_{i}^{C} ( \Psibardot_{i}  + \Phibardot_{i}) \big) \crffM \Phibar_{j}+\right.\\
&~~~~~~~~~~~~~~~~~~~~~~~~~~~~\left.2\Bten{\I_{i}^{C}}{\Phibar_{i}}\Psibardot_{j} \right]\Rten, (k \prec j \preceq i) \\
&   \frac{\partial^2 \taubar_{j}}{\partial \qd_{k} \partial \q_{i}} = \Phibar_{j}\T \big[ 2 \big(   \Bten{\I_{i}^{C}}{\Psibardot_{i}}+ \red{\mathcal{A}_{1}} \big) \Phibar_{k} + \blue{\mathcal{A}_{2}}   (\Psibardot_{k} + \Phibardot_{k} ) \big],\\ &~~~~~~~~~~~~~~~~~~~~~~~~~~ (k \preceq j \prec i) \\
&  \frac{\partial^2 \taubar_{k}}{\partial \qd_{j} \partial \q_{i}} =  \Phibar_{k}\T \Big[2 \big(   \Bten{\I_{i}^{C}}{\Psibardot_{i}}+\red{\mathcal{A}_{1}} \big) \Phibar_{j}+  \blue{\mathcal{A}_{2}}(\Psibardot_{j} + \Phibardot_{j} )  \Big], \\ &~~~~~~~~~~~~~~~~~~~~~~~~~~ (k \prec j \prec i)
\end{align*}
}

\noindent \underline {\bf FO Partials of $\M(\q)$ w.r.t $\q$:}
{\small
\begin{align*}
    \small
&      \frac{\partial \M_{ji}}{\partial \q_{k}}  =  \frac{\partial \M_{ij}}{\partial \q_{k}}= 0, (k \preceq j \preceq i)\\
&        \frac{\partial \M_{ki}}{\partial \q_{j}} =  \Phibar_{k}\T ((\I_{i}^{C}\Phibar_{i}) \crffM \Phibar_{j})\Rten   ,  (k \prec j \preceq i) \\
&  \frac{\partial \M_{ik}}{\partial \q_{j}} = \left[\frac{\partial \M_{ki}}{\partial \q_{j}}  \right]\Tten ,  (k \prec j \preceq i) \\
&      \frac{\partial \M_{kj}}{\partial \q_{i}} = \Phibar_{k}\T  \blue{\mathcal{A}_{2}} \Phibar_{j}, (k \preceq  j \prec i) \\
&     \frac{\partial \M_{jk}}{\partial \q_{i}} = \left[ \frac{\partial \M_{kj}}{\partial \q_{i}} \right ]\Tten  ,  (k \preceq  j \prec i)
\end{align*}
}


{\normalsize
\begin{algorithm*}[t]
\normalsize
\caption{IDSVA-SO Algorithm. {\em Temporary variables in the algorithm are sized as $\A_i \in \mathbb{R}^{6 \times 6}$ for matrices, and  $\u_i \in \mathbb{R}^{6\times 1}$ for vectors.}}
\begin{multicols}{2}
\begin{algorithmic}[1]  
\REQUIRE $ \q, \,\qd ,\, \qdd,\, model$
\setlength{\itemindent}{.005cm}
\addtolength{\algorithmicindent}{-.01cm}
\setstretch{1.07} 
\STATE $ \v_0 = 0; \, \a_{{0}} = -\a_g  $

\FOR{$j=1$ to $N$} \label{alg:forward_start}

\STATE  $ \v_{i} =  \v_{\lambda(i)} +  \Phibar_i \qd_{i}$

\STATE $\a_{i} =  \a_{\lambda(i)} +  \Phibar_i \ddot{\q}_{i} + \v_i \times \Phibar_{i} \qd_i$ \\

\STATE $ \Phibardot_i  =  \v_{i} \times \Phibar_i  $\\[.5ex] 

\STATE $\Psibardot_i  =  \v_{\lambda(i)}\times \Phibar_i  $\\[.5ex] 
\STATE $\Psibarddot_i  =  \a_{\lambda(i)}\times \Phibar_i + \v_{\lambda(i)}\times \Psibardot_i   $ \\[.5ex]

\STATE $\I_i^C = \I_i$
\STATE $ \B_i^C =  ( \v_{i} \times^*) \I_i -  \I_i  ( \v_{i} \times ) + (\I_i \v_i )\crff   $\\[.5ex]

\STATE $ \f_i^C =  \I_i  \a_{i} + ( \v_{i} \times^*) \I_i \v_{i}  $\\[.5ex]  

\ENDFOR \label{alg:forward_end}


\FOR{$i=N$ to $1$} \label{alg:1a_start}

\FOR{$p=1$ to $n_{i}$} \label{alg:1b_start}
\STATE $\phibar_{p} = \Phibar_{i,p}; \psibardot_{p} = \Psibardot_{i,p}; \psibarddot_{p}=\Psibarddot_{i,p}; \phibardot_{p}=\Phibardot_{i,p}$\label{alg:1_define}

\STATE $ \Bmat{\I_{i}^{C}}{\phibar_{p}} = ( \phibar_{p} \timesf) \IC{i} -  \IC{i}  ( \phibar_{p} \times ) + (\IC{i} \phibar_{p} )\crff   $\\
\STATE $ \Bmat{\I_{i}^{C}}{\psibardot_{p}} =  ( \psibardot_{p} \timesf) \IC{i} -  \IC{i}  ( \psibardot_{p} \times ) + (\IC{i} \psibardot_{p} )\crff $ \\
\STATE $\A_{0} = (\IC{i}\phibar_{p})\crff  $\\
\STATE $\A_{1} = \phibar_{p}\timesf \IC{i} - \IC{i}\times  \phibar_{p} $\\
\STATE $\A_{2} = 2 \A_{0} - \Bmat{\I_{i}^{C}}{\phibar_{p}} $\\
\STATE $\A_{3} =  \Bmat{\I_{i}^{C}}{\psibardot_{p}} + \phibar_{p}\timesf \BC{i} - \BC{i}\times  \phibar_{p} $\\
\STATE $\A_{4} = (\BCT{i}\phibar_{p}) \crff $\\
\STATE $\A_{5} = (\BC{i}\psibardot_{p} + \IC{i}\psibarddot_{p}+\phibar_{p}\timesf\f_{i})\crff $\\
\STATE $\A_{6} = \phibar_{p}\timesf \IC{i} + \A_{0}$ \\
\STATE $\A_{7} = (\BC{i}\phibar_{p} + \IC{i}(\psibardot_{p}+\phibardot_{p}))\crff$

\STATE $j=i$
 \WHILE{$j>0$} \label{alg:2a_start}  
 \FOR{$t=1$ to $n_{j}$} \label{alg:2b_start}
\STATE $\phibar_{t} = \Phibar_{j,t}; \psibardot_{t} = \Psibardot_{j,t}; \psibarddot_{t}=\Psibarddot_{j,t}; \phibardot_{t}=\Phibardot_{j,t}$\label{alg:2_define}

  \STATE $\u_{1} = \A_{3}\T\phibar_{t} ; \: \u_{2} =  \A_{1}\T\phibar_{t}$ 
  \STATE $\u_{3} = \A_{3}\psibardot_{t}+\A_{1} \psibarddot_{t}+\A_{5} \phibar_{t}$
  \STATE $\u_{4} = \A_{6} \phibar_{t}; \: \u_{5} = \A_{2} \psibardot_{t}+ \A_{4} \phibar_{t}$
  \STATE $\u_{6} = \Bmat{\I_{i}^{C}}{\phibar_{p}}\psibardot_{t}+\A_{7} \phibar_{t} $
  \STATE $\u_{7} = \A_{3} \phibar_{t}+\A_{1} (\psibardot_{t}+\phibardot_{t})$
  \STATE $\u_{8} = \A_{4} \phibar_{t}-\Bmat{\I_{i}^{C}}{\phibar_{p}}\T\psibardot{}_{t} ; \; \u_{9} = \A_{0} \phibar_{t}$
  \STATE $\u_{10} = \Bmat{\I_{i}^{C}}{\phibar_{p}}\phibar_{t};\: \u_{11} =\Bmat{\I_{i}^{C}}{\phibar_{p}}\T \phibar_{t} $
  \STATE $ \u_{12} = \A_{1}\phibar_{t}$
  
         \STATE $k=j$

         \WHILE{$k>0$} \label{alg:3a_start} 
          \FOR{$r=1$ to $n_{k}$} \label{alg:3b_start}
            \STATE $\phibar_{r} = \Phibar_{k,r}; \psibardot_{r} = \Psibardot_{k,r}$\label{alg:3_define}
            \STATE $\psibarddot_{r}=\Psibarddot_{k,r}; \phibardot_{r}=\Phibardot_{k,r}$

             \STATE $p_{1} = \u_{11}\T \psibardot_{r}$
             \STATE $p_{2}=\u_{8}\T \psibardot_{r} +\u_{9}\T \psibarddot_{r}$
             \STATE $\dtaudqSO{i,p}{j,t}{k,r}=p_{2}; \: \dtaudqdMSO{i,p}{k,r}{j,t}=-p_{1}$ \\[\mysp]
              \IF{$j \neq i$}
                \STATE$\dtaudqSO{j,t}{k,r}{i,p}=\dtaudqSO{j,t}{i,p}{k,r}=\u_{1}\T \psibardot_{r}+\u_{2}\T \psibarddot_{r}$ \\[\mysp]
                \STATE$\dtaudqdMSO{j,t}{i,p}{k,r}= \u_{1}\T \phibar_{r}+\u_{2}\T (\phibardot_{r}+\psibardot_{r}) $ \\[\mysp]
                \STATE $\dtaudqdMSO{j,t}{k,r}{i,p}=p_{1}$ \\[\mysp]
              \STATE$\dtaudqdSO{j,t}{k,r}{i,p}=\dtaudqdSO{j,t}{i,p}{k,r}=\u_{11}\T \phibar_{r}$ \\[\mysp]
             \STATE$\dMdq{k,r}{j,t}{i,p}=\dMdq{j,t}{k,r}{i,p} = \phibar_{r}\T\u_{12}$  \\[\mysp]                
                
              \ENDIF
              
          
            \IF{$k \neq j$}
              \STATE $\dtaudqSO{i,p}{k,r}{j,t}=p_{2}; \: \dtaudqSO{k,r}{i,p}{j,t}= \phibar_{r}\T\u_{3} $ \\[\mysp]
             \STATE $\dtaudqdSO{i,p}{j,t}{k,r}=\dtaudqdSO{i,p}{k,r}{j,t}=-\u_{11}\T \phibar_{r}$\\[\mysp]
            \STATE $\dtaudqdMSO{i,p}{j,t}{k,r}= \phibar_{r}\T\u_{5}+\u_{9}\T (\phibardot_{r}+\psibardot_{r})$\\[\mysp]   
             \STATE $\dtaudqdMSO{k,r}{j,t}{i,p}=  \phibar_{r}\T\u_{6}$ \\[\mysp] \STATE$\dMdq{k,r}{i,p}{j,t}=\dMdq{i,p}{k,r}{j,t} = \phibar_{r}\T \u_{9} $\\[\mysp]

            \IF{$j \neq i$}
                \STATE $\dtaudqSO{k,r}{j,t}{i,p}=\dtaudqSO{k,r}{i,p}{j,t} $\\[\mysp] 
                \STATE $\dtaudqdMSO{k,r}{i,p}{j,t}=\phibar_{r}\T \u_{7} $\\[\mysp]
                \STATE$\dtaudqdSO{k,r}{i,p}{j,t}=\dtaudqdSO{k,r}{j,t}{i,p}=\phibar_{r}\T\u_{10} $\\[\mysp]
            \ELSE
                 \STATE $\dtaudqdSO{k,r}{j,t}{i,p}= \phibar_{r}\T \u_{4} $
            \ENDIF
            \ELSE
                \STATE $\dtaudqdSO{i,p}{j,t}{k,r}=-\u_{2}\T \phibar_{r}  $
            \ENDIF
    
        \ENDFOR \label{alg:3b_end}

          \STATE $k = \lambda(k)$ 
        
         \ENDWHILE \label{alg:3a_end}
        
     \ENDFOR \label{alg:2b_end}
  \STATE $j = \lambda(j)$ 
 \ENDWHILE \label{alg:2a_end}
    \ENDFOR \label{alg:1b_end}

    \IF {$\lambda(i) > 0$}
        \STATE $\I_{\lambda(i)}^C = \I_{\lambda(i)}^C + \I_i^C ; \, \B_{\lambda(i)}^C = \B_{\lambda(i)}^C +  \B_i^C $ \\[.5ex]
        \STATE $ \f_{\lambda(i)}^C = \f_{\lambda(i)}^C +  \f_i^C $ 
    \ENDIF
\ENDFOR \label{alg:1a_end}

\RETURN $\frac{\partial^{2}\taubar}{\partial \q^{2}},\frac{\partial^{2}\taubar}{\partial \qd^{2}},\frac{\partial^{2}\taubar}{\partial \q \partial \qd}, \frac{\partial \M}{\partial \q}$
\end{algorithmic}
\end{multicols}

\label{alg:tau_SO_v5}
\end{algorithm*}
}



 
%

 \bibliographystyle{IEEEtran}
\bibliography{main.bib}

\begin{thebibliography}{10}
\providecommand{\url}[1]{#1}
\csname url@samestyle\endcsname
\providecommand{\newblock}{\relax}
\providecommand{\bibinfo}[2]{#2}
\providecommand{\BIBentrySTDinterwordspacing}{\spaceskip=0pt\relax}
\providecommand{\BIBentryALTinterwordstretchfactor}{4}
\providecommand{\BIBentryALTinterwordspacing}{\spaceskip=\fontdimen2\font plus
\BIBentryALTinterwordstretchfactor\fontdimen3\font minus
  \fontdimen4\font\relax}
\providecommand{\BIBforeignlanguage}[2]{{%
\expandafter\ifx\csname l@#1\endcsname\relax
\typeout{** WARNING: IEEEtran.bst: No hyphenation pattern has been}%
\typeout{** loaded for the language `#1'. Using the pattern for}%
\typeout{** the default language instead.}%
\else
\language=\csname l@#1\endcsname
\fi
#2}}
\providecommand{\BIBdecl}{\relax}
\BIBdecl

\bibitem{tassa2012synthesis}
Y.~Tassa, T.~Erez, and E.~Todorov, ``Synthesis and stabilization of complex
  behaviors through online trajectory optimization,'' in \emph{IEEE/RSJ Int.
  Conf. on Intelligent Robots and Systems}, 2012, pp. 4906--4913.

\bibitem{koenemann2015whole}
J.~Koenemann, A.~Del~Prete, Y.~Tassa, E.~Todorov, O.~Stasse, M.~Bennewitz, and
  N.~Mansard, ``Whole-body model-predictive control applied to the hrp-2
  humanoid,'' in \emph{IEEE/RSJ Int. Conf. on Intelligent Robots and Systems},
  2015, pp. 3346--3351.

\bibitem{dantec2021whole}
E.~Dantec, R.~Budhiraja, A.~Roig, T.~Lembono, G.~Saurel, O.~Stasse,
  P.~Fernbach, S.~Tonneau, S.~Vijayakumar, S.~Calinon \emph{et~al.}, ``Whole
  body model predictive control with a memory of motion: Experiments on a
  torque-controlled talos,'' in \emph{2021 IEEE International Conference on
  Robotics and Automation (ICRA)}.\hskip 1em plus 0.5em minus 0.4em\relax IEEE,
  2021, pp. 8202--8208.

\bibitem{mayne1966second}
D.~Mayne, ``A second-order gradient method for determining optimal trajectories
  of non-linear discrete-time systems,'' \emph{Int. J. of Control}, vol.~3,
  no.~1, pp. 85--95, 1966.

\bibitem{tassa2014control}
Y.~Tassa, N.~Mansard, and E.~Todorov, ``Control-limited differential dynamic
  programming,'' in \emph{IEEE Int. Conf. on Robotics and Automation}, 2014,
  pp. 1168--1175.

\bibitem{chatzinikolaidis2021trajectory}
I.~Chatzinikolaidis and Z.~Li, ``Trajectory optimization of contact-rich
  motions using implicit differential dynamic programming,'' \emph{IEEE
  Robotics and Automation Letters}, vol.~6, no.~2, pp. 2626--2633, 2021.

\bibitem{mastalli2020crocoddyl}
{C.~Mastalli et al.}, ``Crocoddyl: An efficient and versatile framework for
  multi-contact optimal control,'' in \emph{IEEE Int. Conf. on Robotics and
  Automation}, 2020, pp. 2536--2542.

\bibitem{li2020hybrid}
H.~Li and P.~M. Wensing, ``Hybrid systems differential dynamic programming for
  whole-body motion planning of legged robots,'' \emph{IEEE Robotics and
  Automation Letters}, vol.~5, no.~4, pp. 5448--5455, 2020.

\bibitem{pellegrini2020multiple}
E.~Pellegrini and R.~P. Russell, ``A multiple-shooting differential dynamic
  programming algorithm. part 1: Theory,'' \emph{Acta Astronautica}, vol. 170,
  pp. 686--700, 2020.

\bibitem{plancher2018performance}
B.~Plancher and S.~Kuindersma, ``A performance analysis of parallel
  differential dynamic programming on a {GPU},'' in \emph{Int. Workshop on the
  Algorithmic Foundations of Robotics}, 2018, pp. 656--672.

\bibitem{wright1999numerical}
S.~Wright, J.~Nocedal \emph{et~al.}, ``Numerical optimization,'' \emph{Springer
  Science}, vol.~35, no. 67-68, p.~7, 1999.

\bibitem{lee2005newton}
S.-H. Lee, J.~Kim, F.~C. Park, M.~Kim, and J.~E. Bobrow, ``Newton-type
  algorithms for dynamics-based robot movement optimization,'' \emph{IEEE
  Transactions on Robotics}, vol.~21, no.~4, pp. 657--667, 2005.

\bibitem{nganga2021accelerating}
J.~N. Nganga and P.~M. Wensing, ``Accelerating second-order differential
  dynamic programming for rigid-body systems,'' \emph{IEEE Robotics and
  Automation Letters}, vol.~6, no.~4, pp. 7659--7666, 2021.

\bibitem{Featherstone08}
R.~Featherstone, \emph{Rigid Body Dynamics Algorithms}.\hskip 1em plus 0.5em
  minus 0.4em\relax Springer, 2008.

\bibitem{squire1998using}
W.~Squire and G.~Trapp, ``Using complex variables to estimate derivatives of
  real functions,'' \emph{SIAM review}, vol.~40, no.~1, pp. 110--112, 1998.

\bibitem{lantoine2012using}
G.~Lantoine, R.~P. Russell, and T.~Dargent, ``Using multicomplex variables for
  automatic computation of high-order derivatives,'' \emph{ACM Transactions on
  Mathematical Software (TOMS)}, vol.~38, no.~3, pp. 1--21, 2012.

\bibitem{cossette2020complex}
C.~C. Cossette, A.~Walsh, and J.~R. Forbes, ``The complex-step derivative
  approximation on matrix lie groups,'' \emph{IEEE Robotics and Automation
  Letters}, vol.~5, no.~2, pp. 906--913, 2020.

\bibitem{hascoet2013tapenade}
L.~Hascoet and V.~Pascual, ``The tapenade automatic differentiation tool:
  principles, model, and specification,'' \emph{ACM Transactions on
  Mathematical Software (TOMS)}, vol.~39, no.~3, pp. 1--43, 2013.

\bibitem{andersson2012casadi}
J.~Andersson, J.~{\AA}kesson, and M.~Diehl, ``Casadi: A symbolic package for
  automatic differentiation and optimal control,'' in \emph{Recent advances in
  algorithmic differentiation}.\hskip 1em plus 0.5em minus 0.4em\relax
  Springer, 2012, pp. 297--307.

\bibitem{walther2009getting}
A.~Walther and A.~Griewank, ``Getting started with adol-c.''
  \emph{Combinatorial scientific computing}, vol. 181, p. 202, 2009.

\bibitem{makino2006cosy}
K.~Makino and M.~Berz, ``Cosy infinity version 9,'' \emph{Nuclear Instruments
  and Methods in Physics Research Section A: Accelerators, Spectrometers,
  Detectors and Associated Equipment}, vol. 558, no.~1, pp. 346--350, 2006.

\bibitem{tedrake2019drake}
R.~Tedrake \emph{et~al.}, ``Drake: Model-based design and verification for
  robotics,'' \emph{URL https://drake. mit. edu}, 2019.

\bibitem{giftthaler2017automatic}
M.~Giftthaler, M.~Neunert, M.~St{\"a}uble, M.~Frigerio, C.~Semini, and
  J.~Buchli, ``Automatic differentiation of rigid body dynamics for optimal
  control and estimation,'' \emph{Advanced Robotics}, vol.~31, no.~22, pp.
  1225--1237, 2017.

\bibitem{bell2012cppad}
B.~M. Bell, ``Cppad: a package for c++ algorithmic differentiation,''
  \emph{Computational Infrastructure for Operations Research}, vol.~57, no.~10,
  2012.

\bibitem{johannessen2019robot}
L.~M.~G. Johannessen, M.~H. Arbo, and J.~T. Gravdahl, ``Robot dynamics with
  urdf \& casadi,'' in \emph{2019 7th International Conference on Control,
  Mechatronics and Automation (ICCMA)}.\hskip 1em plus 0.5em minus 0.4em\relax
  IEEE, 2019, pp. 1--6.

\bibitem{astudillo2021mixed}
A.~Astudillo, J.~Carpentier, J.~Gillis, G.~Pipeleers, and J.~Swevers, ``Mixed
  use of analytical derivatives and algorithmic differentiation for nmpc of
  robot manipulators,'' \emph{IFAC-PapersOnLine}, vol.~54, no.~20, pp. 78--83,
  2021.

\bibitem{carpentier2018analytical}
J.~Carpentier and N.~Mansard, ``Analytical derivatives of rigid body dynamics
  algorithms,'' in \emph{Robotics: Science and systems}, 2018.

\bibitem{kudruss2019efficient}
M.~Kudruss, P.~Manns, and C.~Kirches, ``Efficient derivative evaluation for
  rigid-body dynamics based on recursive algorithms subject to kinematic and
  loop constraints,'' \emph{IEEE Control Systems Letters}, vol.~3, no.~3, pp.
  619--624, 2019.

\bibitem{mastalli2022inverse}
C.~Mastalli, S.~P. Chhatoi, T.~Corb{\`e}res, S.~Tonneau, and S.~Vijayakumar,
  ``Inverse-dynamics mpc via nullspace resolution,'' \emph{arXiv preprint
  arXiv:2209.05375}, 2022.

\bibitem{jain1993linearization}
A.~Jain and G.~Rodriguez, ``Linearization of manipulator dynamics using spatial
  operators,'' \emph{IEEE transactions on Systems, Man, and Cybernetics},
  vol.~23, no.~1, pp. 239--248, 1993.

\bibitem{ayusawa2018comprehensive}
K.~Ayusawa and E.~Yoshida, ``Comprehensive theory of differential kinematics
  and dynamics towards extensive motion optimization framework,'' \emph{Int. J.
  of Robotics Research}, vol.~37, no. 13-14, pp. 1554--1572, 2018.

\bibitem{singh2022efficient}
S.~Singh, R.~Russell, and P.~M. Wensing, ``Efficient analytical derivatives of
  rigid-body dynamics using spatial vector algebra,'' \emph{IEEE Robotics and
  Automation Letters}, vol.~7, no.~2, pp. 1776--1783, 2022.

\bibitem{bos2022}
\BIBentryALTinterwordspacing
M.~Bos, S.~Traversaro, D.~Pucci, and A.~Saccon, ``Efficient geometric
  linearization of moving-base rigid robot dynamics,'' \emph{Journal of
  Geometric Mechanics}, vol.~14, no.~4, pp. 507--543, 2022. [Online].
  Available: \url{/article/id/6317f7f54cedfd00073d0105}
\BIBentrySTDinterwordspacing

\bibitem{singh2022analytical}
S.~Singh, R.~P. Russell, and P.~M. Wensing, ``Analytical second-order partial
  derivatives of rigid-body inverse dynamics,'' in \emph{2022 IEEE/RSJ
  International Conference on Intelligent Robots and Systems (IROS)}.\hskip 1em
  plus 0.5em minus 0.4em\relax IEEE, 2022, pp. 11\,781--11\,788.

\bibitem{orin1979kinematic}
D.~E. Orin, R.~McGhee, M.~Vukobratovi{\'c}, and G.~Hartoch, ``Kinematic and
  kinetic analysis of open-chain linkages utilizing {Newton-Euler} methods,''
  \emph{Math. Biosciences}, vol.~43, no. 1-2, pp. 107--130, 1979.

\bibitem{vereshchagin1974computer}
A.~Vereshchagin, ``Computer simulation of the dynamics of complicated
  mechanisms of robot-manipulators,'' \emph{Eng. Cybernet.}, vol.~12, pp.
  65--70, 1974.

\bibitem{featherstone1983calculation}
R.~Featherstone, ``The calculation of robot dynamics using articulated-body
  inertias,'' \emph{The international journal of robotics research}, vol.~2,
  no.~1, pp. 13--30, 1983.

\bibitem{echeandia2021numerical}
S.~Echeandia and P.~M. Wensing, ``Numerical methods to compute the coriolis
  matrix and christoffel symbols for rigid-body systems,'' \emph{Journal of
  Comp. and Nonlinear Dynamics}, vol.~16, no.~9, 2021.

\bibitem{park2018geometric}
F.~C. Park, B.~Kim, C.~Jang, and J.~Hong, ``Geometric algorithms for robot
  dynamics: A tutorial review,'' \emph{Applied Mechanics Reviews}, vol.~70,
  no.~1, 2018.

\bibitem{traversaro2019multibody}
S.~Traversaro and A.~Saccon, ``Multibody dynamics notation (version 2),'' 2019.

\bibitem{murray2017mathematical}
R.~M. Murray, Z.~Li, and S.~S. Sastry, \emph{A mathematical introduction to
  robotic manipulation}.\hskip 1em plus 0.5em minus 0.4em\relax CRC press,
  2017.

\bibitem{chirikjian2011stochastic}
G.~S. Chirikjian, \emph{Stochastic models, information theory, and Lie groups,
  volume 2: Analytic methods and modern applications}.\hskip 1em plus 0.5em
  minus 0.4em\relax Springer Science \& Business Media, 2011, vol.~2.

\bibitem{matlab_prop_unit}
S.~Singh and P.~M. Wensing,
  \url{https://github.com/shubhamsingh91/spatial_v2_extended/blob/main/v3/unit_tests/UnitTest_SVATensor_prop.m},
  2023.

\bibitem{garofalo2013closed}
G.~Garofalo, C.~Ott, and A.~Albu-Sch{\"a}ffer, ``On the closed form computation
  of the dynamic matrices and their differentiations,'' in \emph{IEEE/RSJ Int.
  Conf. on Intelligent Robots and Systems}, 2013, pp. 2364--2359.

\bibitem{arxivss_SO}
S.~Singh, R.~P. Russell, and P.~M. Wensing, ``Details of second-order partial
  derivatives of rigid-body inverse dynamics,'' 2022, arXiv:2203.00679.

\bibitem{matlab_iden_unit}
S.~Singh and P.~M. Wensing,
  \url{https://github.com/shubhamsingh91/spatial_v2_extended/blob/main/v3/unit_tests/UnitTest_SVATensor_iden.m},
  2023.

\bibitem{matlab_SO_unit}
------,
  \url{https://github.com/shubhamsingh91/spatial_v2_extended/blob/main/v3/unit_tests/UnitTest_IDSO_expressions.m},
  2023.

\bibitem{matlabsource}
------,
  \url{https://github.com/ROAM-Lab-ND/spatial_v2_extended/blob/main/v3/derivatives/ID_SO_derivatives.m},
  2022, see commit: b06fd78, 03/01/2022.

\bibitem{singh2022closed}
S.~Singh, R.~P. Russell, and P.~M. Wensing, ``Closed-form second-order partial
  derivatives of rigid-body inverse dynamics,'' \emph{arXiv preprint
  arXiv:2203.01497}, 2022.

\bibitem{carpentier2019pinocchio}
{J.~Carpentier et al.}, ``The {Pinocchio} {C++} library: A fast and flexible
  implementation of rigid body dynamics algorithms and their analytical
  derivatives,'' in \emph{IEEE/SICE Int. Symposium on System Integration},
  2019, pp. 614--619.

\bibitem{cppsource}
S.~Singh,
  \url{https://github.com/shubhamsingh91/pinocchio/blob/master/src/algorithm/rnea_SO_derivatives.hxx},
  2022.

\bibitem{carpentier2018analyticalMinv}
J.~Carpentier, ``Analytical inverse of the joint space inertia matrix,'' 2018,
  available: https://hal.laas.fr/hal-01790934.

\bibitem{plancher2021accelerating}
B.~Plancher, S.~M. Neuman, T.~Bourgeat, S.~Kuindersma, S.~Devadas, and V.~J.
  Reddi, ``Accelerating robot dynamics gradients on a cpu, gpu, and fpga,''
  \emph{IEEE Robotics and Automation Letters}, vol.~6, no.~2, pp. 2335--2342,
  2021.

\end{thebibliography}

\vfill

\end{document}